\documentclass[11pt,a4paper]{article}
\usepackage{authblk}
\usepackage[hyperref]{acl2021}
\usepackage{times,balance}
\usepackage{latexsym}

\usepackage{todonotes}
\usepackage{amsmath,comment}
\usepackage{tikz}
\usepackage{subfig,enumitem}
\usepackage{multirow}
\usepackage{multicol}
\usepackage{cleveref}
\usepackage{xcolor}

\aclfinalcopy

\def\checkmark{\tikz\fill[scale=0.2](0,.35) -- (.25,0) -- (1,.7) -- (.25,.15) -- cycle;}

\title{Fingerprinting Fine-tuned Language Models in the Wild\thanks{Supplementary contains dataset, source code, and an appendix (including hyper-parameter setting and additional results). The code and dataset are available at \href{https://github.com/LCS2-IIITD/ACL-FFLM}{\nolinkurl{https://github.com/LCS2-IIITD/ACL-FFLM}} .}  }

\author[1]{Nirav Diwan}
\author[1]{Tanmoy Chakravorty}
\author[2]{Zubair Shafiq}

\affil[1]{IIIT-Delhi, India}
\affil[2]{University of California, Davis}

\affil[1]{\texttt{\{nirav17072,tanmoy\}@iiitd.ac.in}}
\affil[2]{\texttt{zubair@ucdavis.edu}}

\begin{document}
\maketitle

\begin{abstract}
There are concerns that the ability of language models (LMs) to generate high quality synthetic text can be misused to launch spam,  disinformation, or propaganda. 
Therefore, the research community is actively working on developing approaches to detect whether a given text is organic or synthetic. 
While this is a useful first step, it is important to be able to further fingerprint the author LM to attribute its origin. 
Prior work on fingerprinting LMs is limited to attributing synthetic text generated by a handful (usually $<10$) of pre-trained LMs. 
However, LMs such as GPT2 are commonly fine-tuned in a myriad of ways (e.g., on a domain-specific text corpus) before being used to generate synthetic text. 
It is challenging to fingerprinting fine-tuned LMs because the universe of fine-tuned LMs is much larger in realistic scenarios.  
To address this challenge, we study the problem of large-scale fingerprinting of fine-tuned LMs in the wild. 
Using a real-world dataset of synthetic text generated by 108 different fine-tuned LMs, we conduct comprehensive experiments to demonstrate the limitations of existing fingerprinting approaches. 
Our results show that fine-tuning itself is the most effective in attributing the synthetic text generated by fine-tuned LMs. 
\end{abstract}

\section{Introduction}

\textbf{Background \& motivation.}
State-of-the-art language models (LMs) can now generate long, coherent, and grammatically valid synthetic text \cite{devlin2019bert,radford2018improving, radford2019language,brown2020language}.
On one hand, the ability to generate high quality synthetic text offers a fast and inexpensive alternative to otherwise labor-intensive useful applications such as summarization and chat bots \cite{yoo2020intelligent,yu2020financial,wang2019text,liu2019text}.  
On the other hand, such high quality synthetic text can also be misused by bad actors to launch spam, disinformation, or propaganda. 
For example, LMs such as Grover \cite{zellers2019defending} are shown to be capable of generating full-blown news articles, from just brief headlines, which are more believable than equivalent human written news articles.
In fact, prior work has shown that humans cannot distinguish between organic (i.e., human written) and synthetic (i.e., generated by LM) text \cite{ippolito2020automatic,jawahar2020automatic,munir-etal-2021-looking}.
Thus, this ability to generate high quality synthetic text can further be misused for social impersonation and phishing attacks because users can be easily misled about the authorship of the text.

\begin{figure}[!t]
    \includegraphics[width=1\columnwidth]{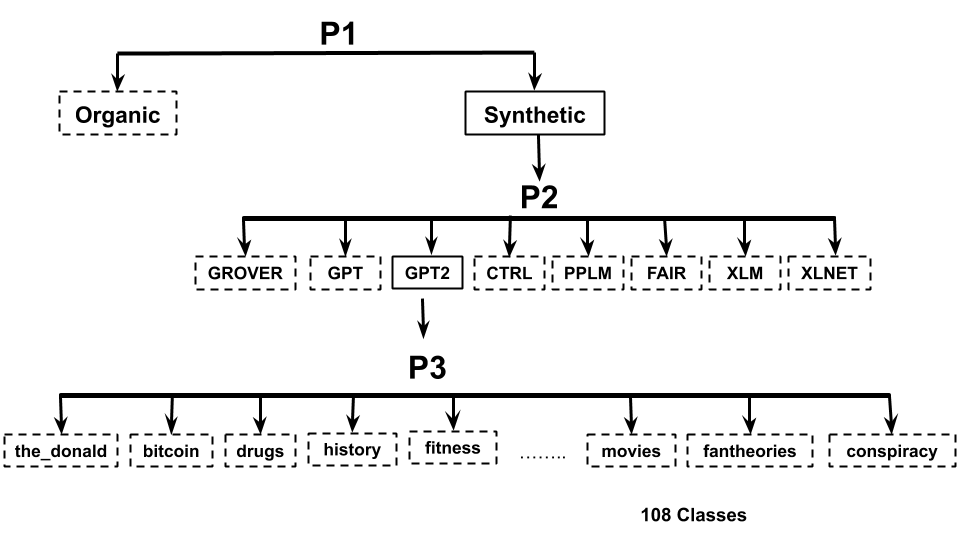}
    \caption{Attribution of text formulated as a series of three problems: P1, P2, and P3.}
    \label{fig:problem_tree}
\end{figure}

\noindent \textbf{Problem statement.}
To mitigate the potential misuse of LMs, the research community has started developing new text attribution techniques. 
However, as shown in Figure \ref{fig:problem_tree}, the attribution of synthetic text is a multistage problem.
The first step is to distinguish between organic and synthetic text (P1). 
Prior work has used the LM's output word probability distribution to detect synthetic text \cite{ippolito2020automatic,gehrmann2019gltr,zellers2019defending}.
However, there are several publicly available pre-trained LMs that might be used to generate synthetic text. 
Thus, the second step is to detect the pre-trained LM used to generate synthetic text (P2). 
Prior approaches showed promising results by attempting to fingerprint the LM based on its distinct semantic embeddings \cite{pan2020privacy,uchendu2020authorship}. 
However, pre-trained LMs such as GPT2 \cite{radford2019language} are commonly fine-tuned before being used to generate synthetic text. 
Thus, the third step is to detect the fine-tuned LM used to generate synthetic text (P3).
To the best of our knowledge, {\em prior work lacks approaches to effectively fingerprint fine-tuned LMs. }

\noindent \textbf{Technical challenges.}
It is particularly challenging to fingerprint fine-tuned LMs simply because of the sheer number of possible fine-tuned variants. 
More specifically, a pre-trained LM can be fine-tuned in a myriad of ways (e.g., separately on different domain-specific text corpora), resulting in a large number of classes. 
Another challenge to fingerprint fine-tuned LMs is that we cannot make assumptions about the nature of fine-tuning (e.g., parameters, training data) or generation (e.g., prompts). 
Prior work on fingerprinting pre-trained LMs is limited to evaluation on a small number of classes ($<10$ classes) and  on synthetic text that is artificially generated using set prompts.

\noindent \textbf{Proposed approach.} To fingerprint synthetic text generated by fine-tuned LMs, we utilize the RoBERTa model \cite{liu2019roberta} and attach a CNN-based classifier on top.
We fine-tune the RoBERTa model for the downstream task of sentence classification using a synthetic text corpus. 
The fine-tuned model is used to extract embeddings as features that are then fed to the CNN classifier. 
We show that the fine-tuned RoBERTa model is able to capture the topic-specific distinguishing patterns of the synthetic text. 
Upon visualizing the generated features, the samples form closely-condensed distinguishable clusters based on the topic of the organic corpus the LMs have been fine-tuned upon. 
Therefore, we conclude that fine-tuning itself significantly helps fingerprinting a fine-tuned LM.
Note that our fingerprinting approach does not assume access to the text corpus used for LM fine-tuning.
We only assume access to arbitrary synthetic text generated by fine-tuned LMs.

\noindent \textbf{Dataset.} We gather a real-world dataset of synthetic text generated by fine-tuned LMs in the wild. 
More specifically, we extract synthetic text from the subreddit r/SubSimulatorGPT2. 
Each of the 108 users on r/SubSimulatorGPT2 is a GPT2 LM that is fine-tuned on 500k posts and comments from a particular subreddit (e.g., r/askmen, r/askreddit,r/askwomen). 
It is noteworthy that users on r/SubSimulatorGPT2 organically interact with each other using the synthetic text in the  preceding comment/reply as their prompt.

\noindent \textbf{Evaluation.} 
We evaluate our models using a suite of evaluation metrics.
We also adapt confidence-based heuristics, such as the gap statistic.
Our best model is accurate for a large number of classes, across a variety of evaluation metrics, showing impressive results for the largest setting of 108 classes. 
While it obtains around 46\% precision and 43\% recall, its top-10 accuracy is about 70\%. 
In other words, the correct class is one of the top-10 predictions in about 70\% of the cases. 
If we give the model the option to not make a classification decision, via confidence estimation, it doubles the precision of the top classification from 46\% to around 87\% with a decrease of recall from 43\% to 27\%.

We summarize our key contributions as follows:

\begin{enumerate}[noitemsep,nolistsep,leftmargin=*]

\item \textbf{Problem formulation.} 
To the best of our knowledge, we are the first to explore the problem of attribution of synthetic text generated by fine-tuned LMs (P3).  
We are also the first to investigate synthetic text attribution for a large number of classes on a real-world dataset. 
We also show that P3 is a much more challenging problem as compared to P1 and P2.

\item \textbf{Comprehensive model.} 
We design and implement a comprehensive set of different feature extraction techniques. 
We use them on a variety of machine learning and deep learning pipelines to build detection models.

\item \textbf{Rigorous evaluation.} 
We conduct rigorous evaluation on a real-world dataset of synthetic text generated by fine-tuned LMs. 
We use several evaluation metrics such as top-k accuracy and precision-recall tradeoff to compare different detection models. 
We also provide insights into the performance of different feature sets and classification algorithms. 
\end{enumerate}

\noindent \textbf{Paper Organization:} The rest of the paper is organized as follows. 
Section \ref{sec: related} contextuailzes our work with respect to prior literature. 
We analyze the real-world dataset of synthetic text generated by fine-tuned LMs in Section \ref{sec: dataset}.
Section \ref{sec: methods} describes different feature sets and classification algorithms for fingerprinting fine-tuned LMs.
Section \ref{sec: results} presents the results of our experimental evaluation  before concluding in Section \ref{sec: conclusion}.

\section{Related Work}
\label{sec: related}
Figure \ref{fig:problem_tree} illustrates three different problem formulations for attribution of synthetic text generated by LMs.
The first line of research (\textbf{P1}) aims to distinguish between organic (by humans) and synthetic (by LMs) text.
Given synthetic text, the second line of research (\textbf{P2}) further aims to attribute the synthetic text generated by pre-trained LMs such as BERT and GPT. 
Finally, in this paper, we further aim to (\textbf{P3}) attribute the synthetic text generated by fine-tuned LMs. 
Here, we first discuss prior work on \textbf{P1} and \textbf{P2} and then highlight the importance and unique challenges of \textbf{P3}.

\textbf{P1:} As the quality of synthetic text generated by LMs has improved, the problem of distinguishing between organic and synthetic text has garnered a lot of attention.  
\citet{gehrmann2019gltr} aimed to distinguish between synthetic text generated by GPT2 and Heliograf versus organic text by books, magazines, and newspapers. 
They showed that humans had a hard time  classifying between organic and synthetic.
Their proposed GLTR model, which uses  probability and ranks of words as predicted by pre-trained LMs as features, was able to achieve 87\% AUC. 
\citet{zellers2019defending} developed Grover, a LM to generate fake news. 
They also showed that humans had a hard time distinguishing between organic and synthetic text generated by Grover.
A classifier based on Grover achieved near perfect accuracy and significantly outperformed other classifiers based on pre-trained LMs. 
\citet{ippolito2020automatic} presented an interesting trade-off between distinguishability of organic and synthetic text.
They showed that synthetic text optimized to fool humans is actually much easily detected by automated classification approaches.
They generated synthetic text from pre-trained GPT2 using different sampling strategies and parameters, and used different classifiers such as GLTR that use pre-trained LMs and a purpose-built fine-tuned BERT based classifier. 
They showed that their fine-tuned BERT based classifier was able to significantly outperform other approaches as well as humans.

\textbf{P2:} 
Given synthetic text, recent work has further attempted to attribute authorship of synthetic text generated by  pre-trained LMs.
\citet{uchendu2020authorship} fingerprinted pre-trained LMs by attributing synthetic text to its author LM. 
They conducted exhaustive experiments using conventional authorship attribution models \cite{kimconvolutional,zhang2015character,cho2014learning} for eight pre-trained LMs -- GPT \cite{radford2018improving}, GPT2 \cite{radford2019language} , GROVER \cite{zellers2019defending}, FAIR \cite{ng2019facebook}, CTRL \cite{keskar2019ctrl}, XLM \cite{lample2019cross}, XLNET \cite{yangxlnet}, and  PPLM \cite{dathathri2019plug}. 
They showed that  derived linguistic features when used with simple classifiers (Random Forest, SVM) perform the best. 
\citet{pan2020privacy} prepared a corpus of organic text and queried each of the five LMs -- BERT \cite{devlin2019bert}, RoBERTa \cite{liu2019roberta}, GPT, GPT2 and XLNET, to generate pre-trained embeddings. 
Then, they trained a multilayer perceptron using the embeddings and obtained perfect accuracy in fingerprinting the pre-trained LM.
\citet{munir-etal-2021-looking} used stylometric features as well as static and dynamic embeddings using ML classifiers to attribute synthetic text generated by four pre-trained LMs -- GPT, GPT2, XLNet, and BART \cite{lewis-etal-2020-bart}.
They obtained near perfect accuracy in fingerprinting pre-trained LMs on a purpose-built dataset of synthetic text.

\textbf{P3:} 
In this paper, we further attempt to attribute authorship of synthetic text generated by fine-tuned LMs.
This problem is relevant in the real-world because malicious actors typically fine-tune a pre-trained LM for domain adaption (e.g., to generate fake news articles vs. food reviews)  \cite{zellers2019defending}.
There are two main novel aspects of   \textbf{P3} that make it more challenging than \textbf{P1} and \textbf{P2}.
{\em First}, LMs can be fine-tuned in numerous different ways. 
More specifically, an attacker can use various datasets to fine-tune a pre-trained LM and further set the fine-tuning parameters and epochs in many different ways. 
Therefore, the size of the universe of fine-tuned LMs is expected to be quite large in \textbf{P3}. 
In \textbf{P1}, the problem is a simple binary classification problem (organic vs. synthetic). 
Similarly, in \textbf{P2}, the problem has still limited number classes because only a handful of pre-trained LMs are publicly available (e.g., GPT, GPT2, etc.).
For instance, \citet{munir-etal-2021-looking}, \citet{pan2020privacy}, and \citet{uchendu2020authorship} respectively considered a universe of  four, five, and eight pre-trained LMs.
In contrast, in this paper, we consider 108 fine-tuned LMs. 
{\em Second}, prior work mostly considered attributing authorship of synthetic text generated in a controlled setting. 
For example, synthetic text is often generated by providing a topical prompt  \cite{uchendu2020authorship}.
As another example, the attribution classifier often assumes some information about the training data    \cite{zellers2019defending}.
In contrast, here we consider the problem of attributing synthetic text in an uncontrolled setting assuming no control or knowledge about training or generation phases of  fine-tuning LM.

\begin{figure*}[t]
\captionsetup[subfigure]{labelformat=empty}
    \centering
    \footnotesize
    \subfloat[\footnotesize{(a) Average \#Words} ]{{\includegraphics[width=.63\columnwidth]{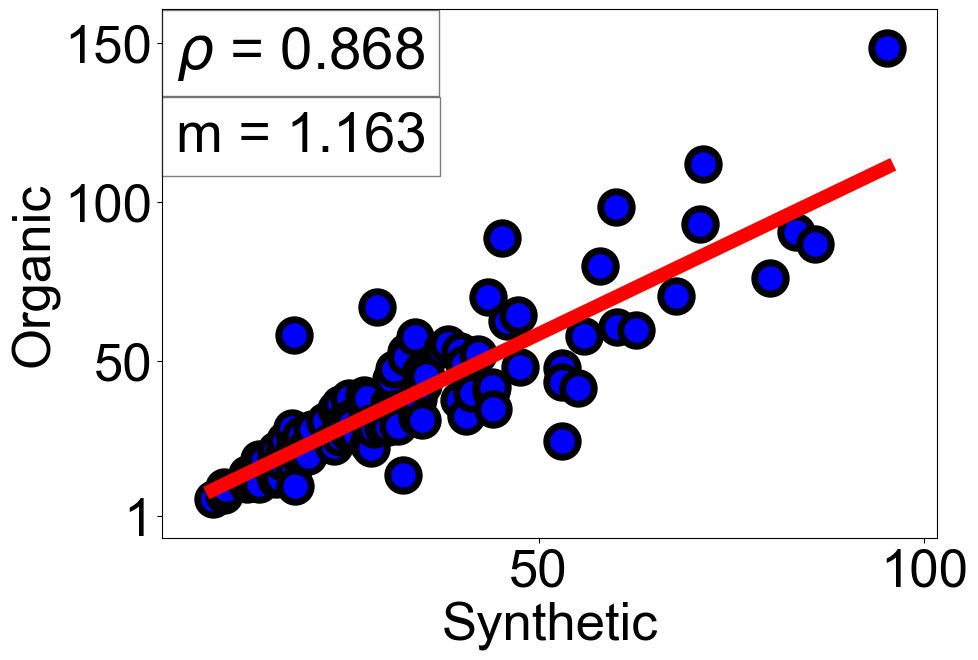}\label{fig:comp_word_avg}}}
    \subfloat[\footnotesize{(b) SD Words} ]{{\includegraphics[width=.63\columnwidth]{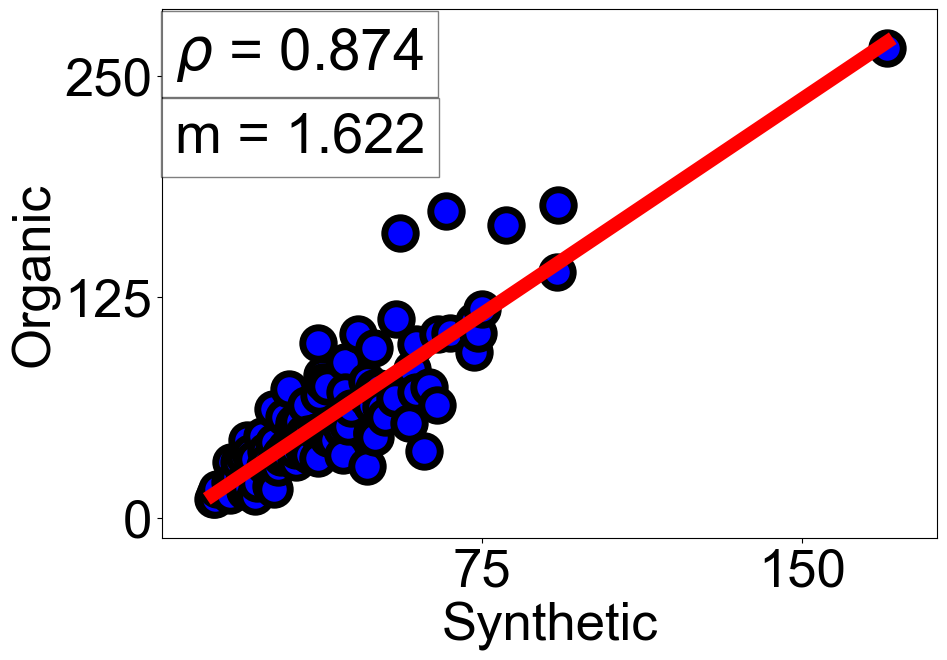}\label{fig:comp_word_sd}}}
    \subfloat[\footnotesize{(c) Average \#Sentences} ]{{\includegraphics[width=.63\columnwidth]{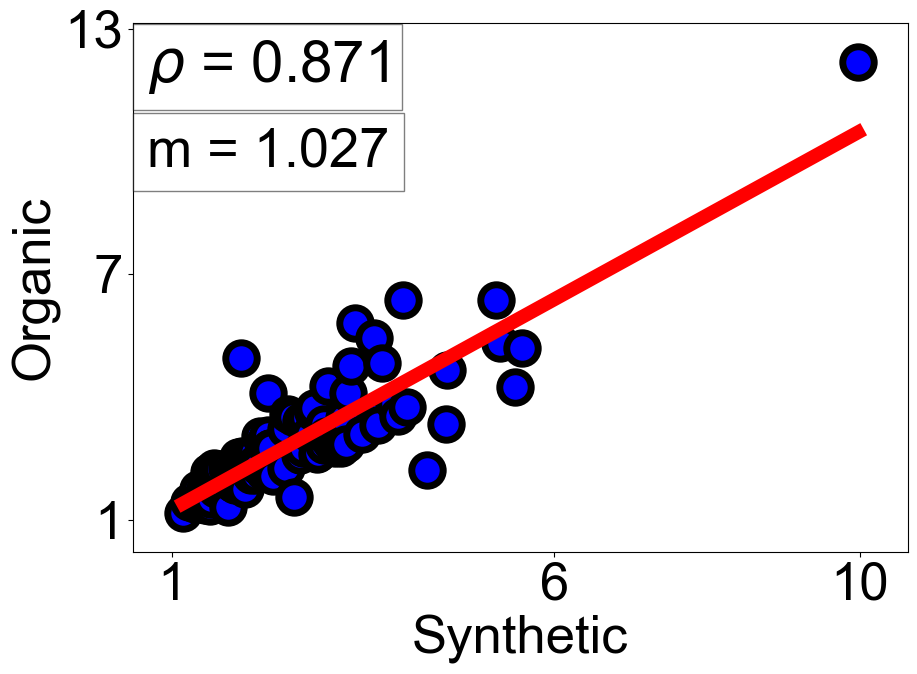}\label{fig:comp_sent_avg}}}
    \vspace{-3mm}
    \subfloat[\footnotesize{(d) SD Sentences} ]{{\includegraphics[width=.63\columnwidth]{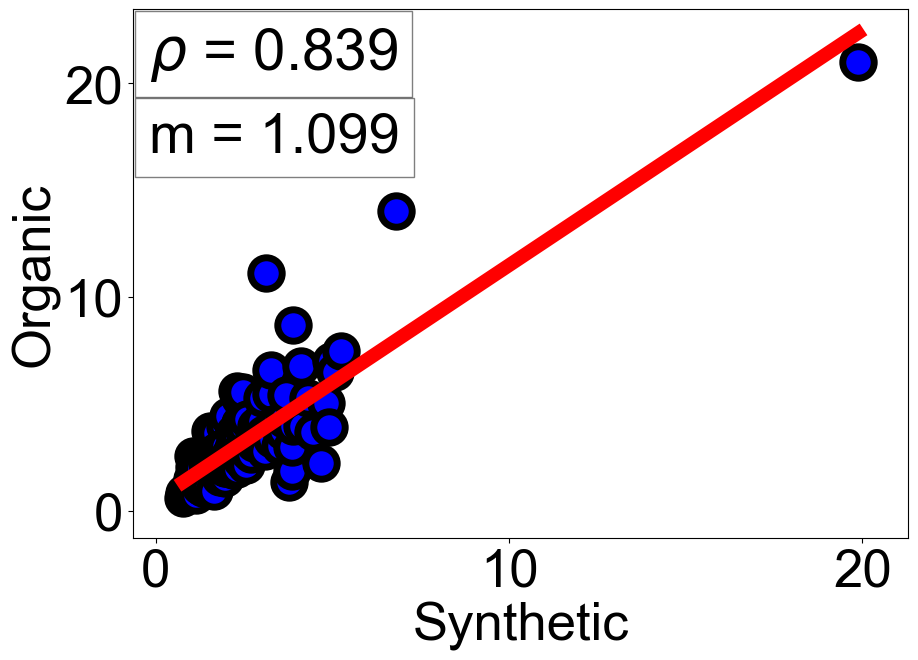}}
    \label{fig:comp_sent_sd}}
    \subfloat[\footnotesize{(e) Vocabulary Size} ]{{\includegraphics[width=.63\columnwidth]{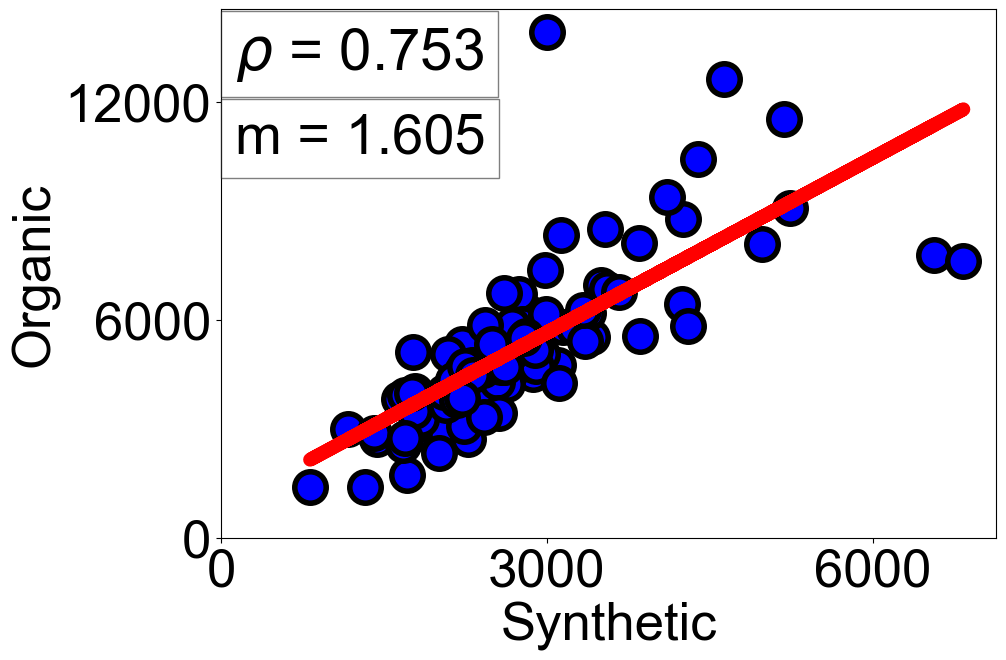} }\label{fig:comp_vocab_size}}
    \subfloat[\footnotesize{(f) KF Readability} ]{{\includegraphics[width=.63\columnwidth]{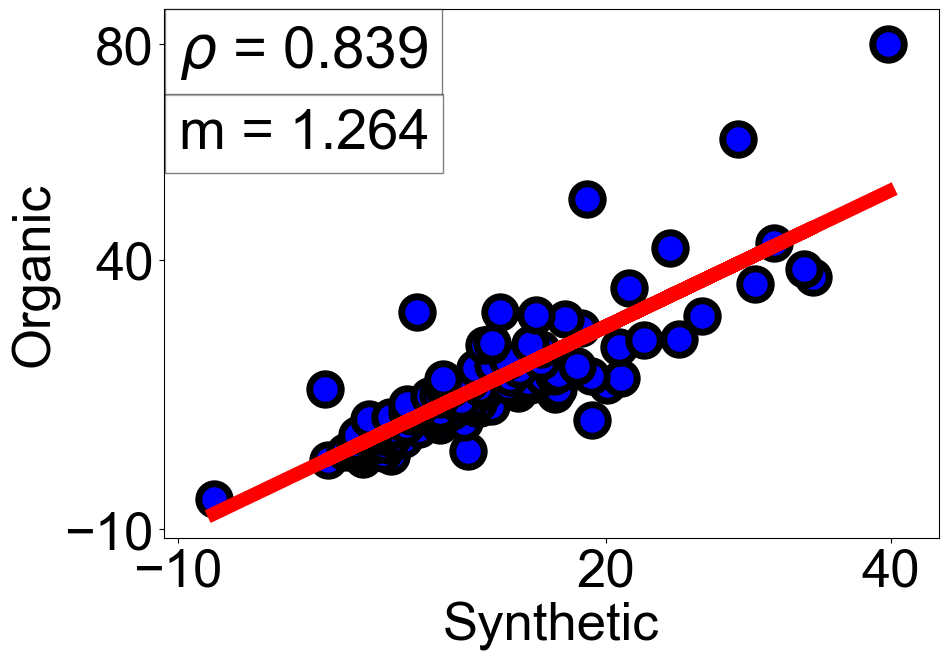}}\label{fig:comp_read}}
    \caption{{A class-wise comparison of organic and synthetic text indicating strong correspondence in terms of: (a) average number of words; (b) standard deviation of number of words; (c) average number of sentences; (d) standard deviation of number of sentences; (e) vocabulary size; and (f) Kincaid-Flescher readability. $\rho$ is the Pearson correlation coefficient and $m$ is the slope of the linear fit.}}
    \label{fig:comp}%
\end{figure*}

\section{Dataset}
\label{sec: dataset}
Prior work on detection and attribution of synthetic text has relied on artificial purpose-built datasets in a controlled environment. 
We overcome this issue by gathering a real-world dataset of synthetic text by different GPT2 bots on the \textsc{r/SubSimulatorGPT2}.
Note that \textsc{r/SubSimulatorGPT2} is independently designed and operated by its moderators.
Each user on the r\textsc{/SubSimulatorGPT2} subreddit is a GPT2 small (345 MB) bot that is fine-tuned on 500k posts and comments from a particular subreddit (e.g., r/askmen, r/askreddit, r/askwomen).
The bots generate posts on r\textsc{/SubSimulatorGPT2}, starting off with the main post followed by comments (and replies) from other bots.
The bots also interact with each other by using the synthetic text in the preceding comment/reply as their prompt. 
In total, the subreddit contains 401,214 comments posted between June 2019 and January 2020 by 108 fine-tuned GPT2 LMs (or class).
The complete details of various design choices are described here: \url{https://www.reddit.com/r/SubSimulatorGPT2Meta/}



\begin{figure*}[t]
\captionsetup[subfigure]{labelformat=empty}
    \centering
    \subfloat[(a) ]{{\includegraphics[width=0.63\columnwidth]{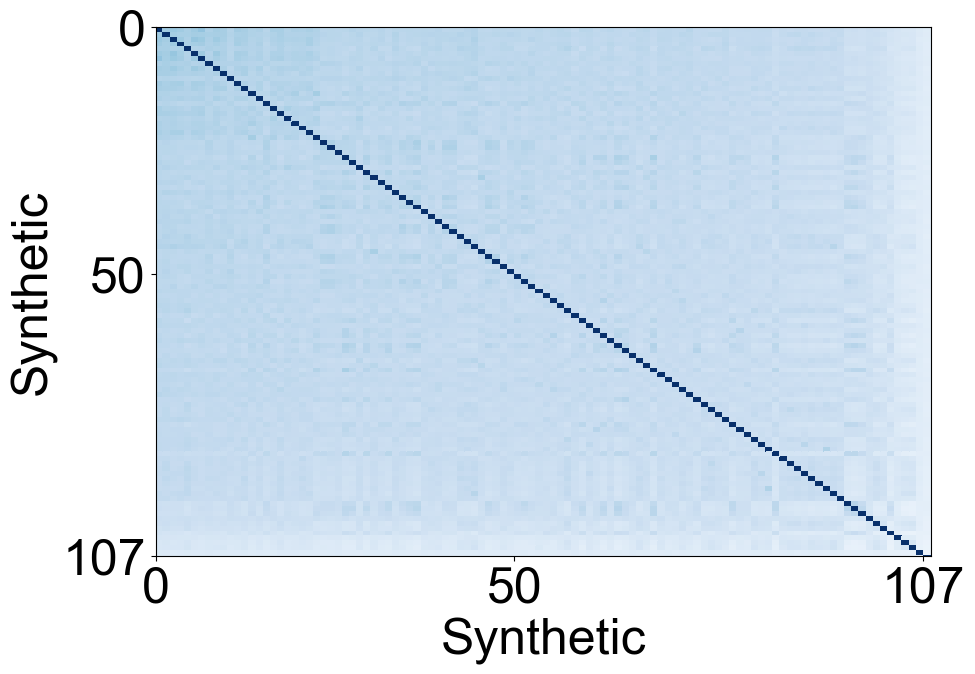} }\label{fig:syn_matrix}}
    \subfloat[(b)]{{\includegraphics[width=0.63\columnwidth]{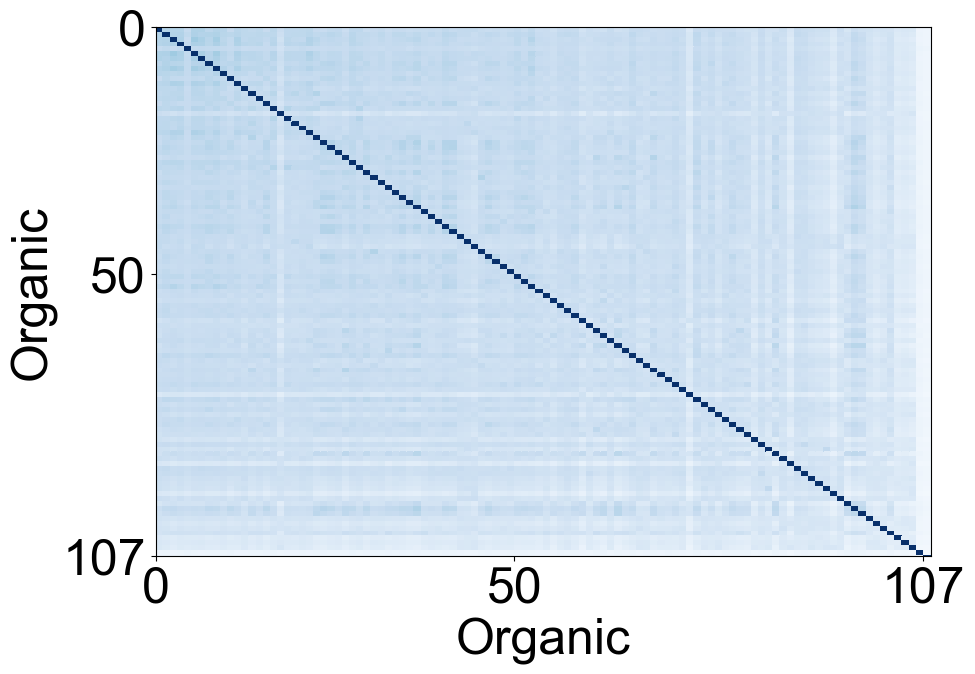}}\label{fig:org_matrix}}
    \subfloat[(c) ]{{\includegraphics[width=0.63\columnwidth]{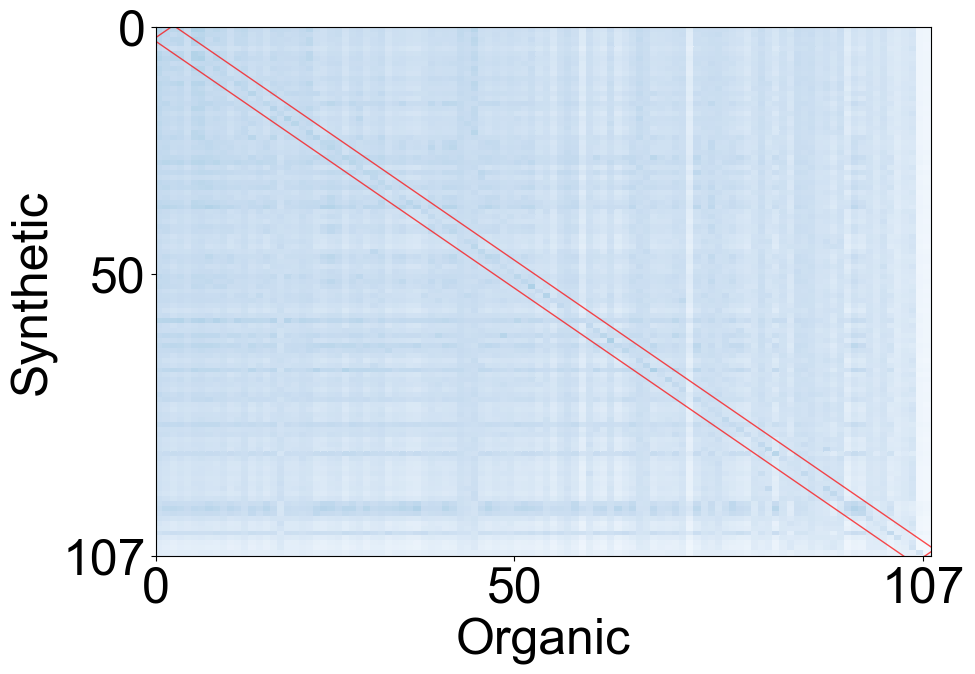}\label{fig:cross_matrix}}}
    \subfloat[]{{\includegraphics[width=0.1\columnwidth, height =3.35cm]{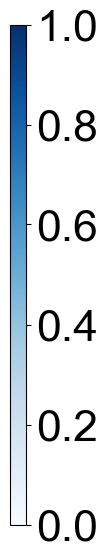}}}
    \caption{{Pair-wise vocabulary overlap between classes of (a) synthetic text, (b) organic text, and (c) cross-comparison of organic and synthetic text. The higher intensity in (a) indicates more vocabulary overlap between classes of (a) synthetic text as compared to (b) organic text. The dark diagonal (highlighted in red box) in (c) indicates  significant vocabulary overlap between the synthetic and organic text from the same subreddit. }}
    \label{fig:matrix}%
\end{figure*}

\begin{table}[!t]
\begin{tabular}{c|c}
 
\footnotesize{\textbf{Subreddit}} &  \footnotesize{\textbf{Synthetic text excerpt}} \\ 
\hline
\footnotesize{{r/conspiracy}} & \footnotesize{\makecell{ I'm sure the elite have been working to\\  control what they do, but I don't think  \\ they have the resources or  manpower   \\to control something as massive as the \\  global economy.}} \\ 
\hline
\footnotesize{{r/confession}} & \footnotesize{\makecell{You need to tell her how you feel! She \\needs to know how you feel so that you\\ can work out how to make the right\\ decision. You can tell her how much\\  you hate being a coward and how you'd \\never be able to live up to your promise.}} \\ 
\hline
\footnotesize{{r/wallstreetbets}} & \footnotesize{\makecell{There's a huge amount of volatility \\right now that  I don't know how well \\  my eyes can handle. I'm not going\\ to buy any puts for the next week.}} \\ 
\hline
\end{tabular}
\caption{Excerpts of synthetic text generated by  GPT2 LMs fine-tuned on different subreddits.}
\label{table:examples}
\end{table}

Table \ref{table:examples} lists some representative examples of the synthetic text generated by three different fine-tuned GPT2 LMs on our dataset.
We note that the synthetic text is fairly coherent and also captures the unique vocabulary and style of the subreddit used for fine-tuning. 
For example, the excerpt from r/conspiracy reads like a conspiracy discussion, the excerpt from r/confession mentions a suggestive reply to the main post, and the excerpt for r/wallstreetbets uses the specialised finance terms like ``puts''.

Next, we quantitatively compare and contrast synthetic and organic texts corresponding to different subreddits. 
To this end, we collect organic text from the  corresponding subreddits. 
Specifically, we randomly sample 1,000 comments from each subreddit class of synthetic and organic text. 
We contrast basic lexical characteristics, vocabulary, and readability of synthetic and organic text.

\textbf{{Lexical.}} 
%
First, we contrast the following basic lexical features: average/standard deviation of number of words/sentences per comment.
We also measure the Pearson correlation coefficient ($\rho$) between pairs of synthetic and organic texts in \Cref{fig:comp_sent_avg,fig:comp_sent_sd,fig:comp_word_avg,fig:comp_word_sd}.
We note a high correlation ($ \rho > 0.83$) across all lexical features. 
Thus, we conclude that there is a strong dependency between the lexical characteristics of the synthetic and organic text.
This finding indicates that {\em synthetic text generated by fine-tuned GPT2 models indeed captures the lexical characteristics of the  corresponding organic text used for fine-tuning}.

\textbf{{Vocabulary.}} 
Second, we compare the vocabularies of synthetic and organic text of each class. 
We do some basic pre-processing: lower-case, tokenize, and lemmatize all words, remove all punctuation and emojis, and replace hyperlinks and numbers with standard tags.
\Cref{fig:comp_vocab_size} compares the vocabulary size of synthetic and organic text.
While organic text seems to have a large vocabulary than synthetic text, we note a strong correlation ($\rho = 0.76$) between their vocabulary sizes.
We further compute Jaccard similarity  to measure pair-wise vocabulary overlap between different synthetic and organic texts.
Figure \ref{fig:matrix} visualizes the similarity matrices between all pairs classes -- synthetic-synthetic, organic-organic, and synthetic-organic. 
It is noteworthy that the left diagonal represents much higher overlap between corresponding pairs of synthetic and organic text classes, even across synthetic-organic text classes.
%
This finding indicates that the {\em fine-tuned GPT2 models indeed pick up the vocabulary from the organic text in the corresponding subreddit}. 
It is also noteworthy that the average vocabulary overlap among synthetic text classes in Figure \ref{fig:syn_matrix} is much higher than that among organic text classes as shown in Figure \ref{fig:org_matrix}.
This  indicates that {\em synthetic text vocabulary combines information from both pre-training and fine-tuning} -- the text corpus used to pre-train the base GPT2 model as well as the text corpus from the particular subreddit used to fine-tune it.

\textbf{Readability.}
Finally, we compare the readability of the synthetic and organic text.
Figure \ref{fig:comp_read} compares the Kincaid-Flescher readability score \cite{kincaid1975derivation} of synthetic and organic text.
We note that the average readability of synthetic text (16.8) is less than that of organic text (11.5). 
This observation corroborates the findings in recent prior work \cite{uchendu2020authorship}.
Similar to lexical and vocabulary analysis, we note a strong correlation ($\rho = 0.84$) between the readability of synthetic and organic text. 


Additionally, using the organic text as reference, we measure the quality of the synthetic text using well-known metrics -- METEOR \cite{banerjee2005meteor} and BLEU \cite{papineni2002bleu}. We observe that the synthetic text achieves high average scores  across all the metrics. We include the results in the appendix.

Overall, we have two key takeaways:
(1) synthetic text is {\em coherent} although with lower overall readability than organic text;
(2) synthetic text captures the {\em characteristics of the organic text used to fine-tune it}.



%

\section{Methods}
\label{sec: methods}
Fingerprinting fine-tuned LMs needs to operate under the following realistic assumptions.  
(i) Fingerprinting methods cannot assume any knowledge about the nature of the LM, fine-tuning (e.g., parameters, layers) or generation (e.g., prompt). 
(ii) They also cannot assume any access to the organic text used for LM fine-tuning. 
(iii) Since a LM can be fine-tuned in a myriad of ways, these methods need to consider a large number of classes.
(iv) These methods are assumed to have access to a limited sample of synthetic text generated by each of the potential fine-tuned LMs.

\subsection{Pre-processing} 
We lower cased the comments and replaced all hyperlinks with a standard tag [\textit{LINK}].
Next, we tokenized the comments. Any comment with 5 or fewer tokens was removed. The maximum number of tokens in a comment is limited to 75. 
In case a comment is larger, only the first 75 tokens in a comment are taken into consideration. This is consistent for all the models we experimented with.
The limit was decided based on the limit of the size of the GPU (11GB) used for fine-tuning GPT2 and RoBERTa models.

\subsection{Features}
We consider several different feature extraction architectures to encode the synthetic text into representation vectors. 
As we discuss later, these representation vectors are then fed to classifiers. 
\textbf{Writeprints:} Writeprints feature set has been widely used for authorship attribution \cite{iqbal2008novel, pearl2012detecting,abbasi2006visualizing}. 
We extract a total of 220 features that include \textit{lexical} (e.g., total number of words, total number of characters, average number of characters per word, digits percentage, upper case characters percentage), \textit{syntactic} (e.g., frequencies of function words, POS tags unigram, bigrams, trigrams), \textit{content-based} (e.g., bag of words, bigrams/trigrams) and \textit{idiosyncratic} (e.g., misspellings percentage) features.

\textbf{GLTR:} \citet{gehrmann2019gltr} used pre-trained LMs  to extract word likelihood features -- word ranks and probabilities. 
We follow the original approach to average the word probabilities of the text based on pre-trained BERT and GPT2.
We also bin the word ranks into 10 unequal ranges. 
The bins are: [1], [2-5], [6-10], [11-25], [26-50], [51-100], [101-250], [251-500], [501-1000], [$>1000$].

\textbf{Glove:} 
Glove embeddings \cite{pennington2014glove} have been commonly used in large-scale authorship attribution \cite{ruder2016character}. 
We follow \cite{kimconvolutional, zhang2015character} to create a representation vector of the size (max \# tokens $\times$ 100, where max \# tokens is set to 75) using Glove. 


\textbf{Pre-trained LMs (GPT2 and RoBERTa):}  
We also extract the embeddings for each comment using the pre-trained GPT2/RoBERTa model. 
Similar to \citet{nils2019sentence,feng2020language, zhu2020sentence},  we take the [CLS] token representation from the last layer to extract the embeddings of size $1 \times 768$. 
The final embeddings are then scaled between the values of [-3, 3]  using min-max normalization.

\textbf{Fine-tuned (FT) LMs (GPT2 and RoBERTa):} 
We add a softmax classification layer to the pre-trained  GPT2/RoBERTa model.
Then, we fine-tune the LM for the task of sentence classification using the synthetic text in the training dataset.
We again extract the embeddings (size = $1\times 768$) by  taking  the [CLS] token representation from the second last layer. 
The final embeddings are then scaled between the values of [-3, 3]  using min-max normalization. 

\subsection{Classifiers}
\hspace{3mm}\textbf{Shallow classifiers:} 
We use a probabilistic classifier -- Gaussian Naive Bayes (GNB), an ensemble decision classifier -- Random Forest (RF), and a feed-forward multilayer perceptron (MLP) across all of the feature representations.

\textbf{CNN classifier:} 
We experiment with a \textit{stacked} CNN model where convolution layers with different kernel  sizes are stacked before the embedding layer \cite{kimconvolutional}. 
%
%
%
%
%
In this architecture, we attach two stacked convolution 1D layer, followed by a batch normalisation layer and a max pooling layer. 
The output is then fed to two dense layers, the latter of which is a softmax layer.
%

In addition, we also experiment with other shallow classifiers (SVM, Decision Tree) and two more types of feature generators (fine-tuned GLTR, trainable word embeddings).
We report additional results, observations and the hyper-parameters for all the models in the appendix.



\section{Evaluation}
\label{sec: results}
We conduct experiments to evaluate these methods using the real-world Reddit dataset as described in Section \ref{sec: dataset}. 
Our training, validation, and test sets consist of 800, 100 and 200 synthetic comments respectively from each of the 108 subreddit classes.
In total, our training, validation and test sets comprise 86k, 11k and 22k comments respectively.
For evaluation, we use macro precision and recall.
We also measure top-$k$ accuracy based on the confidence score to assess the accuracy of the classifiers in $k$ ($k= 5,10$) guesses for 108 classes.

\begin{table}[!t]
\centering
\footnotesize
\resizebox{0.9\columnwidth}{!}{
\begin{tabular}{l|l|r|r|r|r}
\hline
        \multirow{2}{*}{\textbf{Architecture}} & \multirow{2}{*}{\textbf{Classifier}} & \multicolumn{2}{c|}{\textbf{Macro}} & \multicolumn{2}{c}{\textbf{Top-$k$}} \\ \cline{3-6} 
                                   &            & Prec        & Recall     & \multicolumn{1}{c|}{5}     & \multicolumn{1}{c}{10}   \\ \hline
\multirow{3}{*}{GLTR}              & GNB        & 5.5       & 4.4    & 12.9  & 20.9  \\  
                                   & RF         & 7.8       & 6.6    & 12.6  & 19.0  \\  
                                   & MLP        & 3.6       & 6.3    & 15.6  & 23.7  \\ \hline
\multirow{4}{*}{Writeprints}       & GNB        & 8.2       & 5.8    & 14.1  & 21.4  \\  
                                   & RF         & 10.2      & 8.4    & 14.9  & 21.8  \\  
                                   & MLP        & 16.9      & 14.7   & 30.8  & 42.1  \\ \hline
\multirow{4}{*}{GloVE}             & GNB        & 19.2      & 9.3    & 21.9  & 31.2  \\  
                                   & RF         & 20.5      & 16.9   & 27.1  & 36.2  \\   
                                   & MLP        & 29.7      & 27.2   & 44.4  & 54.1  \\ 
                                   & CNN        & 31.1      & 26.7   & 44.2  & 53.5  \\ \hline
\multirow{4}{*}{GPT2}              & GNB        & 24.8      & 12.4  & 27.8 & 37.7 \\  
                                   & RF         & 10.5      & 7.8    & 15.8  & 27.1 \\  
                                   & MLP        & 44.9      & 29.0   & 47.5 & 56.9 \\  
                                   & CNN        & 30.9      & 28.7   & 49.1  & 59.1  \\ \hline
\multirow{4}{*}{RoBERTa}           & GNB        & 39.2      & 15.8   & 30.8  & 41.0  \\  
                                   & RF         & 11.1      & 8.4    & 16.6  & 25.8  \\  
                                   & MLP        & 44.0      & 34.8   & 54.8  & 62.5  \\  
                                   & CNN        & 33.5      & 32.0   & 53.1  & 63.0  \\ \hline
\multirow{5}{*}{FT-GPT2}           & GNB        & 40.1      & 37.0   & 56.9  & 66.0 \\  
                                   & RF         & 27.6      & 22.8   & 34.8  & 45.2 \\  
                                   & MLP        & 40.2      & 36.4   & 55.7  & 64.0 \\  
                                   & CNN        & 44.6      & 42.1   & 60.9  & 68.9 \\ \hline
\multirow{5}{*}{FT-RoBERTa}        & GNB        & \textbf{47.7}      & 41.5  & 57.9  & 64.9  \\ 
                                   & RF         & 42.0      & 36.8   & 46.9  & 53.2  \\  
                                   & MLP        & 42.8      & 41.5   & 58.2  & 65.3  \\  
                                   & CNN        &  46.0     & \textbf{43.6}   & \textbf{62.0}    & \textbf{69.7}  \\ \hline
\end{tabular}
}
\caption{
Performance of multi-class classifiers based on macro Precision (Prec), Recall and top-$k$ accuracy ($k = 5, 10$) for the largest setting of 108 classes. 
}
\label{tab:results}
\end{table}

\subsection{Results}
\begin{figure}[!t]
\captionsetup[subfigure]{labelformat=empty}
    \centering
    \subfloat[ ]{{\includegraphics[width=7.5cm, height=1.0cm]{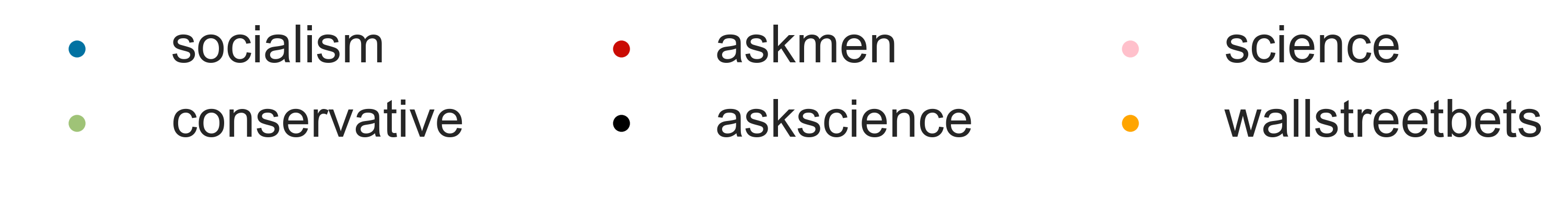}}}
    \vspace{-10mm}
    \subfloat[\footnotesize{(a) Fine-tuned RoBERTa} ]{{\includegraphics[width=3.80cm]{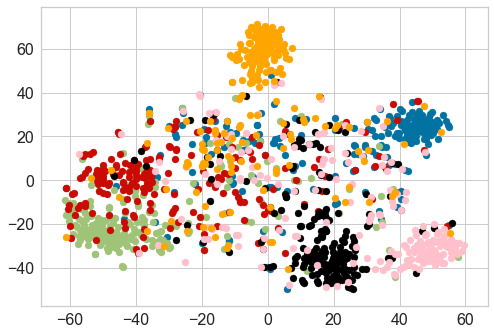}\label{fig:viz_fine_roberta}}}
    \subfloat[\footnotesize{(b) Pre-trained RoBERTa} ]{{\includegraphics[width=3.80cm]{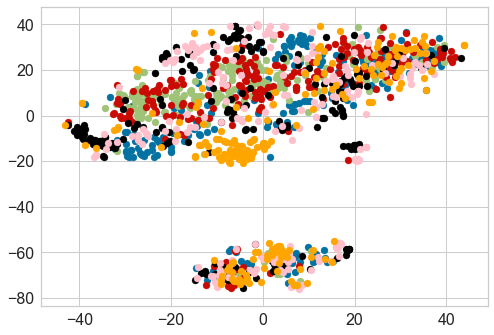}\label{fig:viz_pre}}}
    
    \subfloat[\footnotesize{(c) Fine-tuned GPT2} ]{{\includegraphics[width=3.80cm]{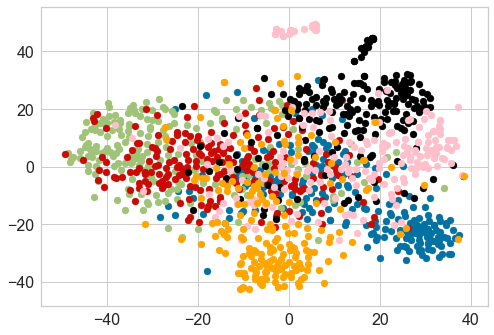}\label{fig:viz_fine_gpt2}}}
    \subfloat[\footnotesize{(d) Pre-trained GPT2} ]{{\includegraphics[width=3.80cm]{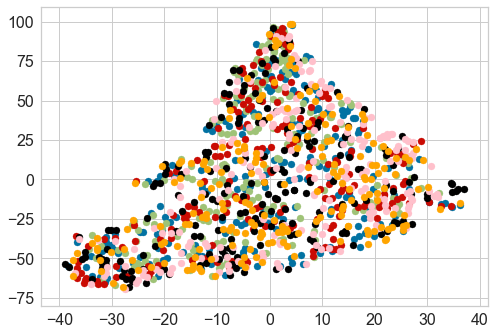}\label{fig:viz_pre_gpt2}}}
    \caption{\footnotesize{Visualisation of  fine-tuned (a,c) and pre-trained embeddings (b,d) of specific classes. Closely condensed clusters specific to the domain of the organic text form in the fine-tuned embeddings.}}
        \label{fig:viz}%
\end{figure}

Table \ref{tab:results} lists the results of different feature representations and classifiers. 
Overall,   classifiers based on fine-tuned LM embeddings perform the best, with  RoBERTa slightly outperforming GPT2.
Fine-tuned  embeddings are successfully able to capture the domain of the organic text the LM is fine-tuned on.
To provide further insights into our best performing feature representations,  we visualize the feature embeddings of pre-trained and fine-tuned RoBERTa and GPT2. 
Figure \ref{fig:viz} plots the 2D projection of synthetic text (using t-SNE) generated by different LMs that are fine-tuned on various subreddits.  
Fine-tuned embeddings form more cohesive and separated clusters than pre-trained embeddings. 
Thus, we conclude that fine-tuning these embeddings is beneficial in attributing synthetic text generated by different fine-tuned LMs. 
Note that certain clusters are more cohesive and better separated than others. 
For example, the most distinct cluster is observed for r/wallstreetbets in Figures \ref{fig:viz}a and \ref{fig:viz}c for fine-tuned embeddings.
However, it is not quite distinct in Figures \ref{fig:viz}b and \ref{fig:viz}d  for pre-trained embeddings. 
We also note that some clusters with high topical similarity (e.g., r/science and r/askscience) are closer to each other. 
On the other hand, some clusters with likely lower topical similarity (e.g., r/socialism and r/conservative) are far apart.

\begin{figure}[]
\captionsetup[subfigure]{labelformat=empty}
    \centering
    \subfloat[\footnotesize{(a) Micro} ]{{\includegraphics[width=0.5\columnwidth]{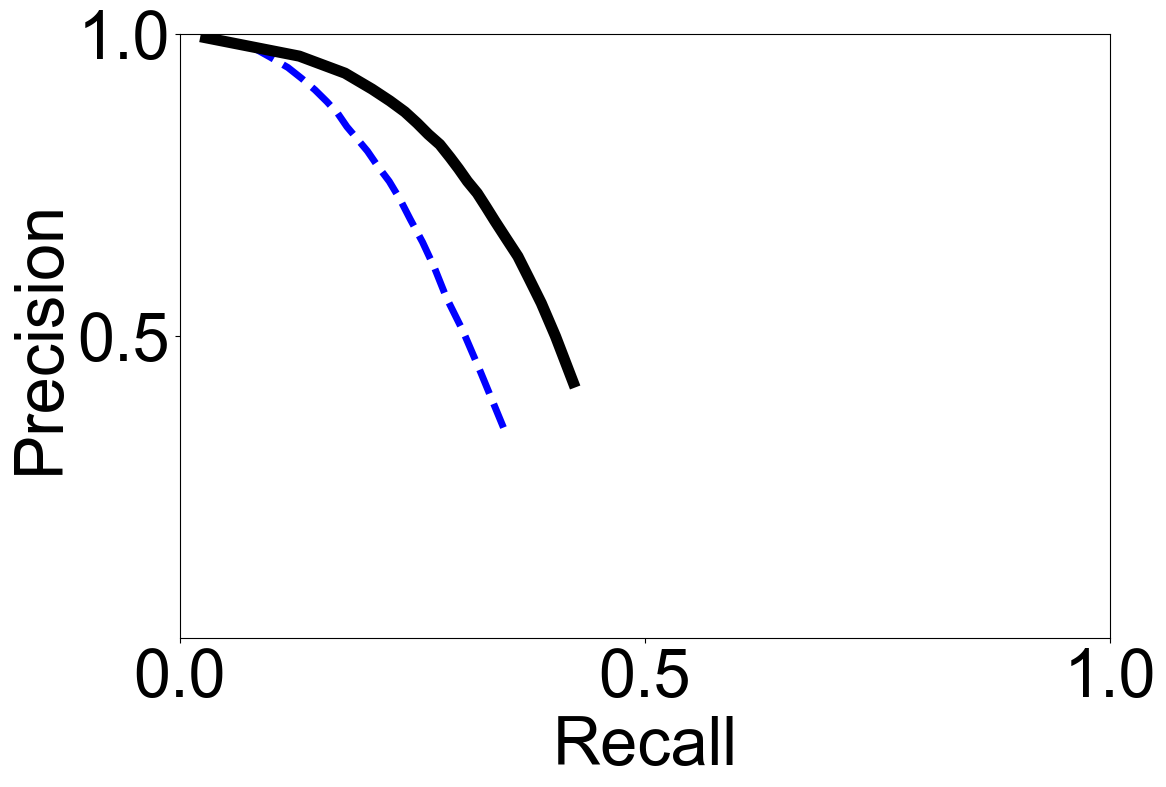}\label{fig:micro}}}
    \subfloat[\footnotesize{(b) Macro} ]{{\includegraphics[width=0.5\columnwidth]{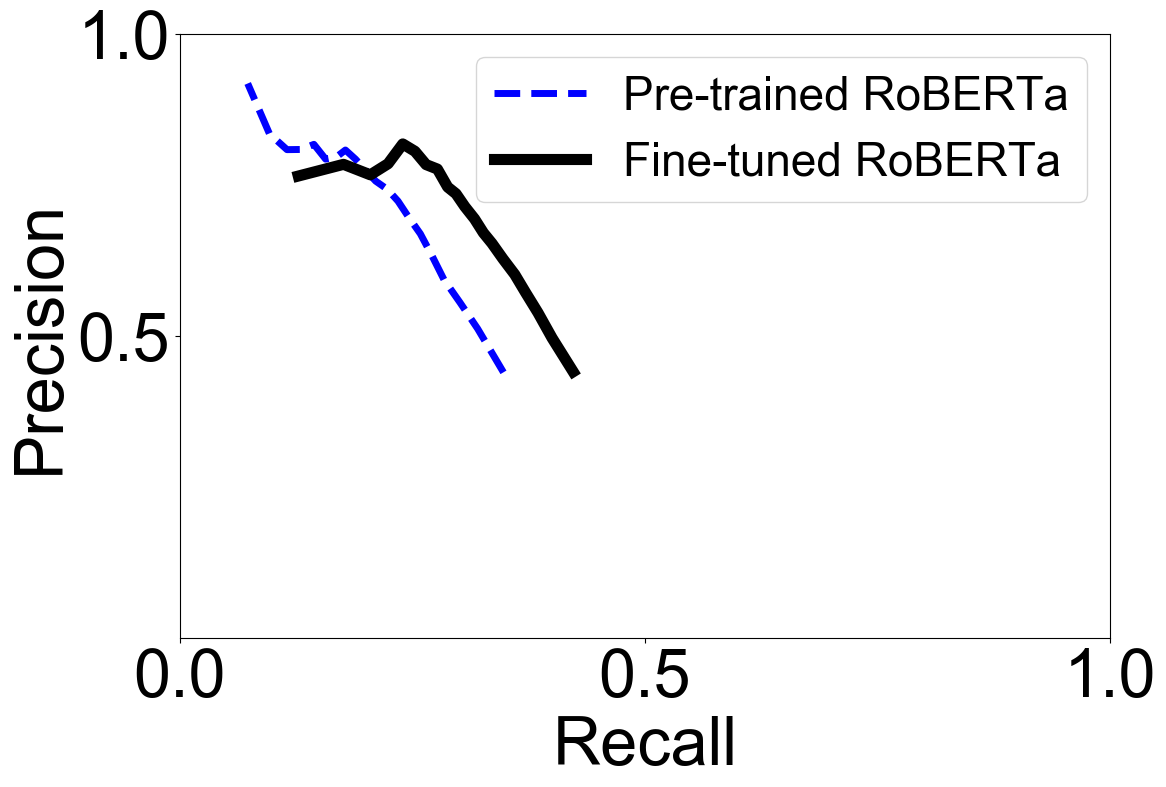}\label{fig:macro}}}
    \caption{\footnotesize{(a) Micro and (b) Macro \textit{precision-recall trade-off} by varying the gap statistic threshold. The comparison with {\em all baselines} is included in the appendix.}}
\end{figure}

\begin{figure*}[!t]
\captionsetup[subfigure]{labelformat=empty}
    \centering
    \subfloat[\footnotesize{(a)} ]{{\includegraphics[width=.5\columnwidth]{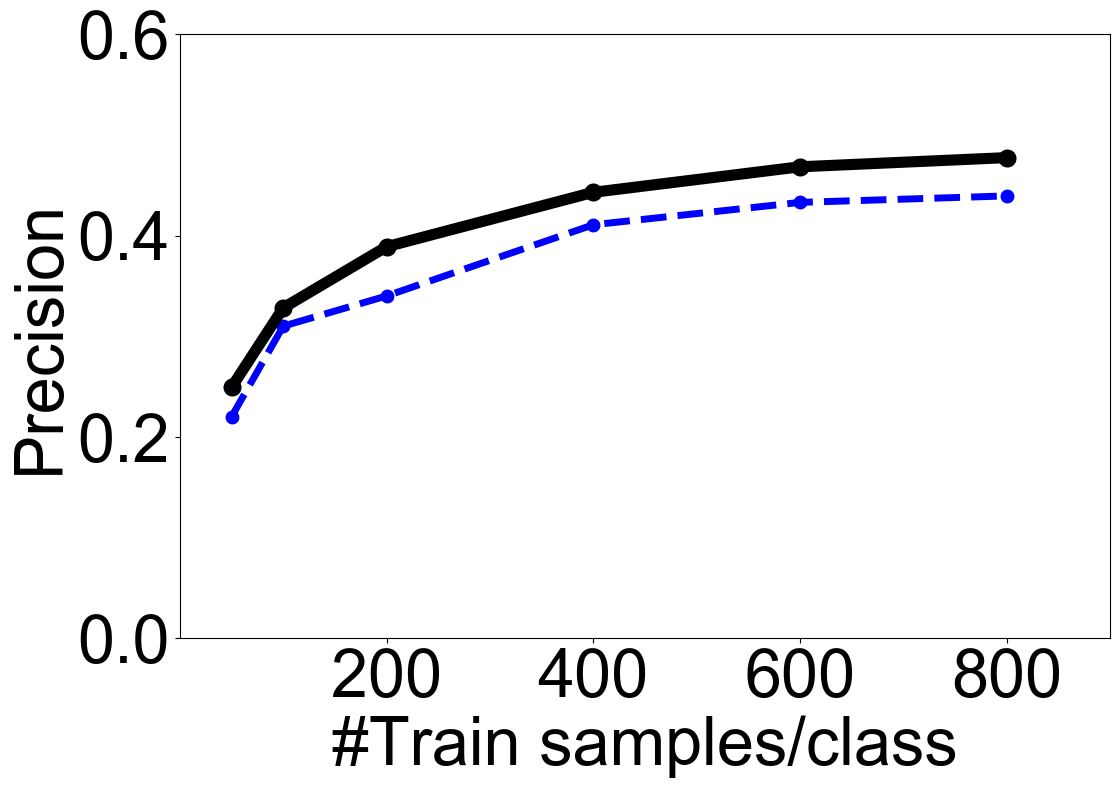}\label{fig:numbertraining_precision}}}
    \subfloat[\footnotesize{(b)} ]{{\includegraphics[width=.5\columnwidth]{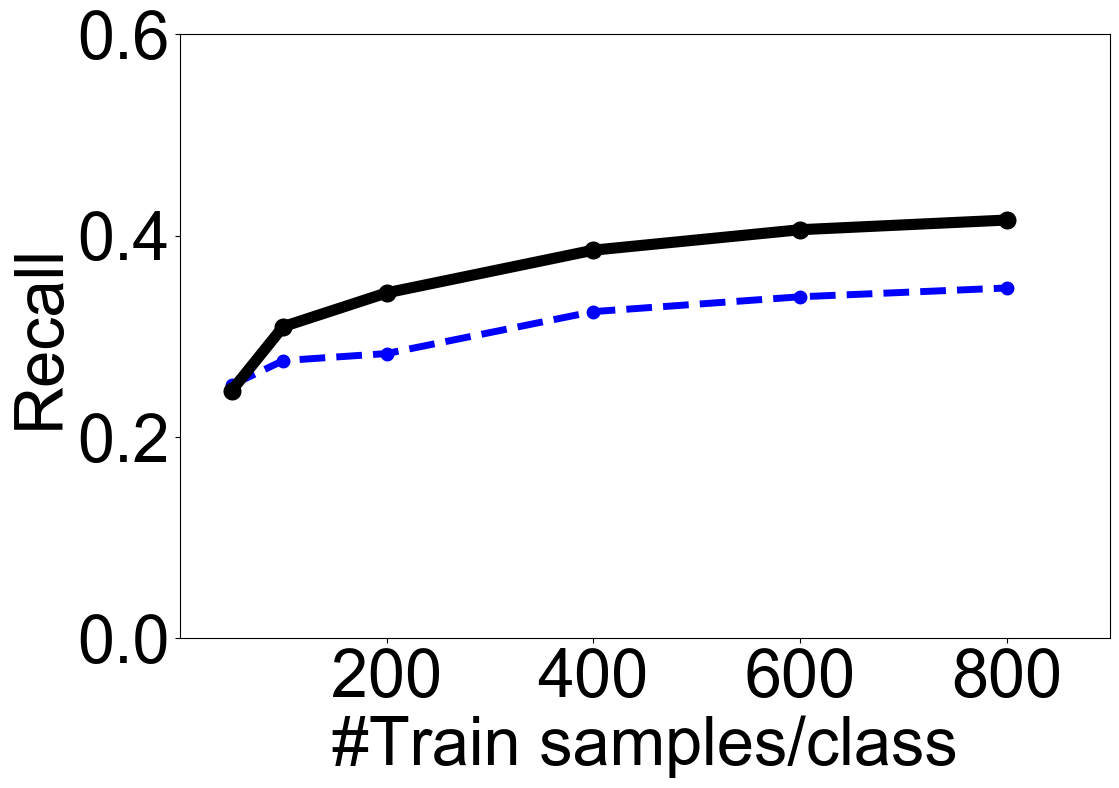}\label{fig:numbertraining_recall}}}
    \subfloat[\footnotesize{(c)} ]{{\includegraphics[width=.5\columnwidth]{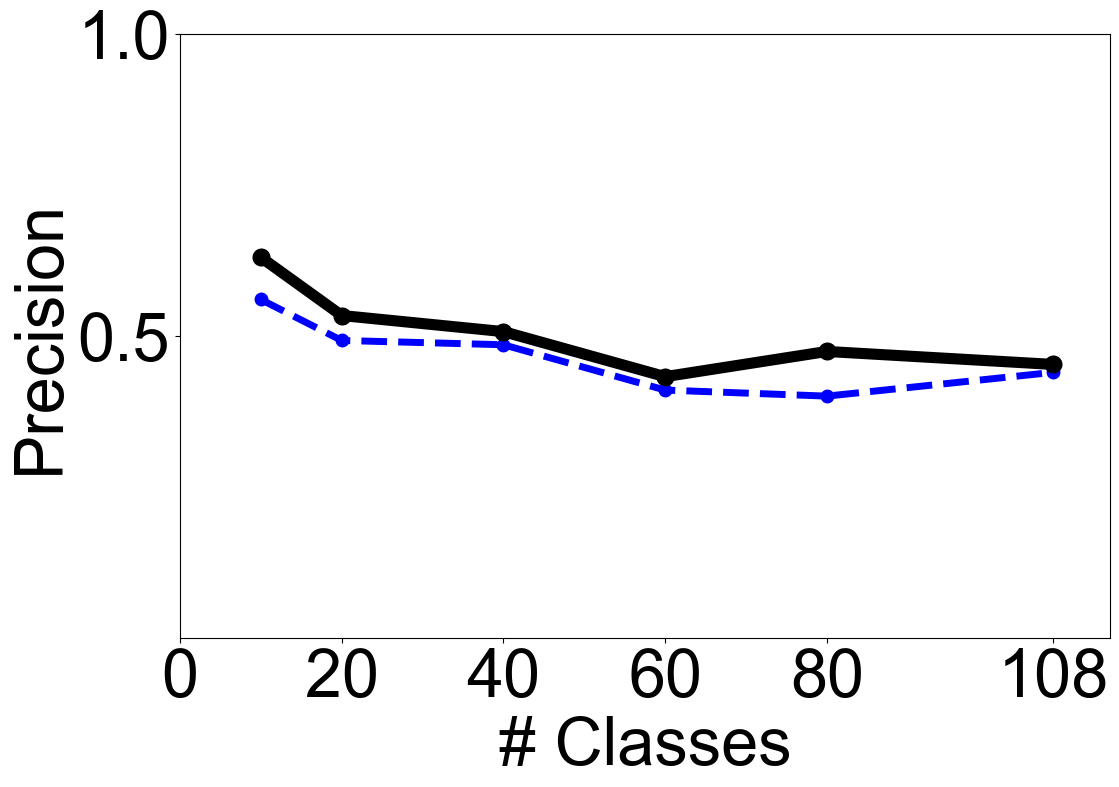}\label{fig:numberclasses_precision}}}
    \subfloat[\footnotesize{(d)} 
    ]{{\includegraphics[width=.5\columnwidth]{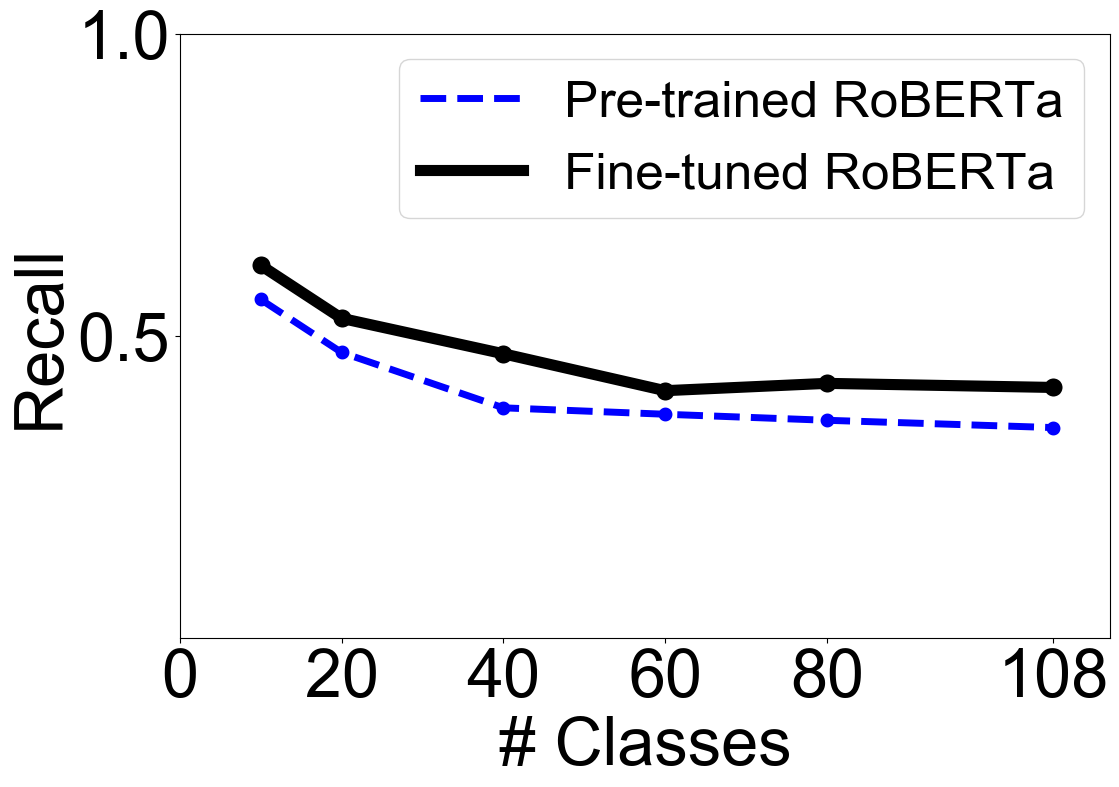}}\label{fig:numberclasses_recall}}
    \caption{\footnotesize{Comparison between the performances of pre-trained and fine-tuned RoBERTa  by varying different parameters. (a) Precision and (b) Recall with the varying training size. (c) Precision and (c) Recall with the varying number of classes. Overall,  fine-tuned RoBERTa   outperforms   pre-trained RoBERTa. The comparison with {\em all baselines} is included in the appendix.}}
\end{figure*}
Despite combining word probabilities from both BERT and GPT2, GLTR is ineffective. 
We find that synthetic texts generated from different fine-tuned models have similar word probabilities  because perhaps they are more impacted by the pre-training process rather than the subsequent fine-tuning.
This shows that the classifier that performs well for  distinguishing between organic and synthetic text (P1) does not work well for  distinguishing between synthetic text by different fine-tuned LMs (P3).
Writeprints feature set provides some improvement but is still ineffective.
Our finding corroborates \citet{manjavacas2017assessing}, who reported that  linguistic and stylistic features as used in Writeprints are not effective in distinguishing between synthetic text.
GloVE again offers some improvement over Writeprints but its performance remains significantly worse than that of our best performing method.

Overall, fine-tuned RoBERTa embeddings with CNN performs the best with  46.0\% precision and 43.6\% recall. 
In about 44\% of the cases, this classifier can correctly fingerprint fine-tuned LM amongst 108 classes; in about 70\% of cases the correct prediction is one of the top-10 guesses.
It is noteworthy that Random Forest which performed exceedingly well in prior work on fingerprinting pre-trianed LMs \cite{uchendu2020authorship}, does not perform well for fingerprinting fine-tuned LMs.
Surprisingly, a relatively simple classifier like GNB achieves comparable precision and recall for our top guess. 
However, CNN outperforms GNB by a small margin, achieving the best top-10 accuracy of 69.7\% for FT-RoBERTa.

\subsection{Discussion}
Next, we analyze the performance of our best performing RoBERTa feature representation and classifier (CNN) under different conditions.\footnote{Other baseline results are reported in the appendix.}

\textbf{Precision-Recall trade-off.}
We evaluate the precision-recall trade-off by imposing a threshold on the confidence of our prediction. 
To this end, we use the \textit{gap statistic},  defined as the difference between the probability of the highest and second highest prediction \cite{narayanan2012feasibility}. 
If the gap statistic is lower than our threshold, the classifier chooses to not make a prediction for the test sample. 
This invariably has an impact on precision and recall.  
Typically, the precision of the classifier increases, since it can more accurately predict the correct class for the samples it has a high confidence in. 
Due to certain samples not being  predicted for, recall is expected to decrease. 
Note that since the classifier may make different number of predictions across classes,  micro and macro precision/recall could be different.

Figures \ref{fig:micro} and \ref{fig:macro} respectively plot the micro and macro precision/recall as we vary the gap statistic.
Overall, the classifier using FT-RoBERTa embeddings achieves a better precision-recall trade-off compared to using standard RoBERTa embeddings. 
As expected, precision improves at the expense of recall for larger values of gap statistic.
Micro precision increases 46\% to over 87\% with a reduction in the micro recall from 43\% to 27\%.  
Similarly, despite potential class imbalance, macro precision increases 46\% to over 81\% with a reduction in the micro recall from 43\% to 26\%.  
Thus, we conclude that the confidence of our best performing fingerprinting method can be tuned to achieve very high precision with some compromise on recall.

\textbf{Impact of number of training samples.} 
Next, we evaluate the impact on the performance of our best models by varying   training size from 50 to 800 samples per class. 
As we vary the  training data, we keep the same test set, i.e., 200 samples from each class. 
 Figures \ref{fig:numbertraining_precision} and \ref{fig:numbertraining_recall}, respectively
 show that precision and recall of FT-RoBERTa plateau at around 400 samples per class. 
Despite at twice the training data, using training 800 samples per class has similar precision/recall as compared to using 400 training samples per class. %
We conclude that having more training samples of synthetic text may not always lead to a significant improvement in fingerprinting performance.

\textbf{Impact of number of classes.} 
We further vary the number of classes from 10 to 108 and report the performance of our best models. 
Figures \ref{fig:numberclasses_precision} and \ref{fig:numberclasses_recall} show that for a 10-class problem, FT-RoBERTa + CNN  achieves 63.0\% precision and 61.7\% recall. 
As expected, both precision and recall decrease as the number of classes increases. 
For 108 classes, the same classifier achieves 46.0\% precision and 43.6\% recall. 
This indicates that fingerprinting a fine-tuned LM is highly challenging  when the universe of potential fine-tuned LMs is larger. 
%

\section{Conclusion}
\label{sec: conclusion}
In this paper, we studied the problem of attribution of synthetic text generated by fine-tuned LMs. 
We designed a comprehensive set of feature extraction techniques and applied them on a number of different machine learning and deep learning pipelines. 
The results showed that the best performing approach used fine-tuned RoBERTa embeddings with CNN classifier.
Our findings present opportunities for future work on fingerprinting LMs in even more challenging open-world scenarios, where the list of potential LMs might be incomplete or synthetic text is not available to train attribution classifiers.

\section*{Ethics and Broader Impact}
Any biases found in the gathered dataset are unintentional, and we do not intend to do harm anyone. 
We would also like to acknowledge that the use of large-scale transformer models requires non-trivial compute resources (GPUs/TPUs) that contributes to global warming \cite{strubell-etal-2019-energy}.
However, we did not train the models from scratch and simply fine-tuned them for our work. 

\section*{Acknowledgement} The work was partially supported by the Ramanujan fellowship and the Infosys Centre for AI, IIIT-Delhi.


\balance
\bibliography{acl2021}
\bibliographystyle{acl_natbib}

\appendix{}
\vfill\eject

\section{Appendix}

\subsection{Implementation, Infrastructure,
Software}
We run all experiments on a 24-core machine  with two Intel(R) Xeon(R) Silver 4116 CPU@2.10GHz CPU's and 512 GB RAM. Additionally, the server has a GeForce RTX 2080 Ti (11 GB) GPU card.  We use huggingface, pytorch (1.4.0) and Tensorflow (v.1.23.0) to implement and evaluate all deep learning models in the paper. For classical ML, we utilize scikit-learn (v.0.23.1). All implementations are done using Python language (v.3.7).

\subsection{Hyper-parameters}

\textbf{Fine-tuned RoBERTa + CNN:} We attach 2 CNN convolutional 1D of kernel sizes 2 and 3 layers back to back. The stride for both the layers is 1. The number of output  filters for each of the two convolutional layers is 16.  We then attach a batch normalization layer of size 16. This is followed by a max pooling layer of size 2 with stride 1. We fine-tuned the model for 15 epochs and monitored the validation loss. If the validation loss did not improve for 5 consecutive epochs, we stopped the fine-tuning. 

\noindent \textbf{Fine-tuned RoBERTa + Dense:} We used RoBERTaForSequenceClassification  wrapper from the huggingface module, which attaches a softmax layer on top of pre-trained RoBERTa model. 
We extracted the embeddings of the second to last layer of size 1x768. For all of our models, we used a \textit{AdamW} optimizer with a learning rate of 0.00005. We used batch size of 48 and only picked the first 75 tokens of each token. We did not use any padding. 


For the soft classifiers, using the fine-tuned  RoBERTa embeddings, we obtained the best results with the following parameters -

\textbf{SVM:} C = 0.1, cachesize=200, classweight = None, coef0 = 0.0, degree=3, gamma=`scale', kernel=`linear', tol=0.001\\
\textbf{MLP:}  activation=`relu', alpha=0.01,epsilon=1e-08, 
hiddenlayersizes = 64, learningrate=`adaptive', learningrateinit=0.0001\\
\textbf{RF} - criterion=`entropy', maxdepth=None, maxfeatures=`auto', maxleafnodes=None, minsamplesleaf = 1\\
\textbf{DT} - criterion = `entropy', maxdepth = None, maxfeatures = None, maxleafnodes=None, minsamplesleaf = 1\\
\textbf{GNB} - Default sklearn parameters performed the best

\begin{figure}[t]
    \centering
    \includegraphics[width=0.75\columnwidth]{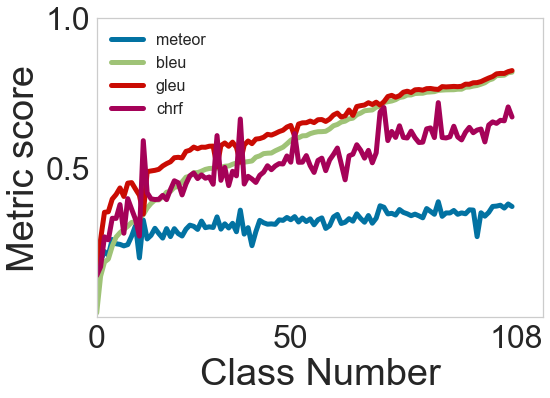}
    \caption{Comparison of different  metrics on the synthetic corpus taking the reference of the organic corpus. The classes  are sorted by increasing CHRF scores.}
    \label{fig:metrics_comp_appendix}
\end{figure}
%


\subsection{Running Time}
All fine-tuned models took a maximum of 8 minutes per epoch. For the soft classifiers, MLP and SVM took the maximum time. Due to the large size of the dataset and the large embedding space, SVM took about 6 hours to train. MLP took an average of 3 hours. Rest of the classifiers took less than an hour for training.

\subsection{Additional Results}
\subsubsection{Additional Dataset Analysis}
Besides the analaysis in the dataset section of the main text, we used 4 metrics - METEOR, BLEU, GLEU and CHRF for  measuring the coherency of the synthetic text, using the organic text as  reference. Using 1000 comments from each class of synthetic and organic text, we obtained  high class-wise average scores on all metrics  - METEOR (0.31), BLEU (0.58), GLEU (0.63) and CHRF (0.51). This provides further evidence that synthetic text is coherent and readable when compared to its organic counterpart. In figure \ref{fig:metrics_comp_appendix} , we illustrate the scores for each individual class, sorted by increasing CHRF scores.

\begin{table}[]
\centering
\resizebox{1\columnwidth}{!}{\begin{tabular}{|l|l|r|r|r|r|}
\hline
\multirow{2}{*}{\textbf{Base Architecture}} & \multirow{2}{*}{\textbf{Classifier}} & \multicolumn{2}{c|}{\textbf{Macro}} & \multirow{2}{*}{\textbf{Top - 5}}& \multirow{2}{*}{\textbf{Top - 10}}\\ \cline{3-4}
                                      &       & \textbf{Precision} & \textbf{Recall} &      &      \\ \hline
\multirow{2}{*}{\textbf{Writeprints}}     & DT    & 6.9       & 6.6    & 6.6  & 6.6  \\ \cline{2-6} 
                                      & SVM   & 19.3      & 16.3   & 33.2 & 44.8 \\ \hline
\multirow{2}{*}{\textbf{GLTR}}                  & DT    & 5.7       & 5.5    & 5.57 & 5.58 \\ \cline{2-6} 
                                      & SVM   & 7.3       & 7.0     & 10   & 13.1 \\ \hline
\multirow{2}{*}{\textbf{RoBERTa}}   & DT    & 6.3       & 5.3    & 6.3  & 6.3  \\ \cline{2-6} 
                                      & SVM   & 8.3       & 6.9    & 7.8  & 8.9  \\ \hline
\multirow{2}{*}{\textbf{Trainable-Word}} & MLP   & 20.4      & 18.9   & 34.3 & 44.3 \\ \cline{2-6} 
                                      & CNN   & 28.6      & 27.0     & 44.0   & 53.1 \\ \hline
\multirow{3}{*}{\textbf{FT RoBERTa}}      & Dense & 44.0       & 42.3   & 60.8 & 68.9 \\ \cline{2-6} 
                                      & DT    & 29.5      & 28.7   & 28.7 & 28.7 \\ \cline{2-6} 
                                      & SVM   & 42.7      & 41.1   & 58.3 & 65.6 \\ \hline

\end{tabular}}
\caption{Results of Decision Tree (DT) and SVM  with the embeddings. We also include the results of the trainable word embeddings model.}
\label{tab:add_results}
\end{table}

\begin{figure*}[t]
\captionsetup[subfigure]{labelformat=empty}
    \centering
    \subfloat[\footnotesize{(a)} ]{{\includegraphics[width=.5\columnwidth]{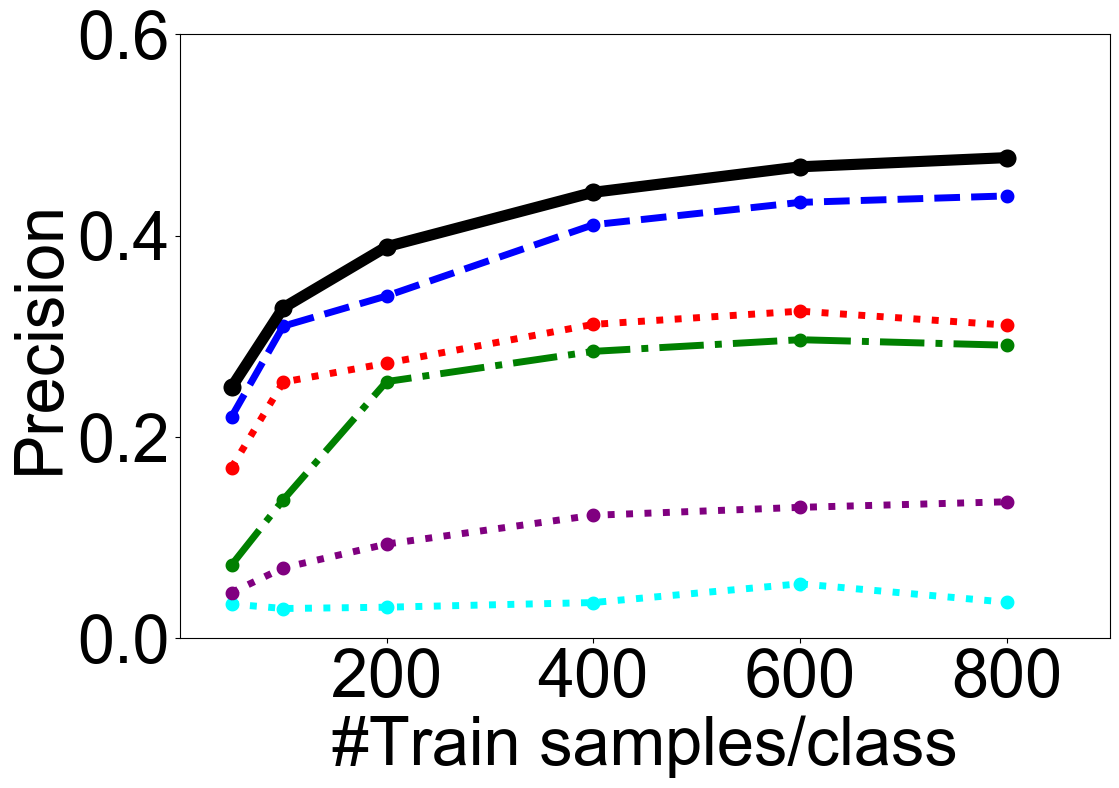}\label{fig:numbertraining_precision}}}
    \subfloat[\footnotesize{(b)} ]{{\includegraphics[width=.5\columnwidth]{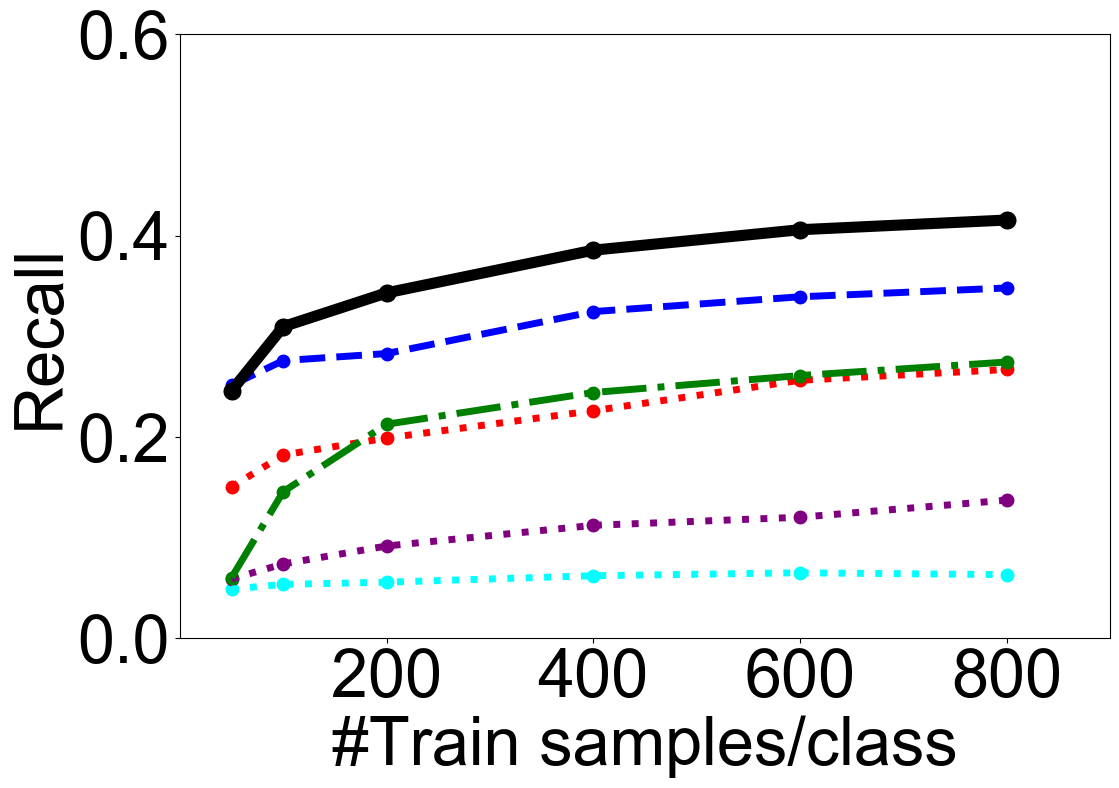}\label{fig:numbertraining_recall}}}
    \subfloat[\footnotesize{(c)} ]{{\includegraphics[width=.5\columnwidth]{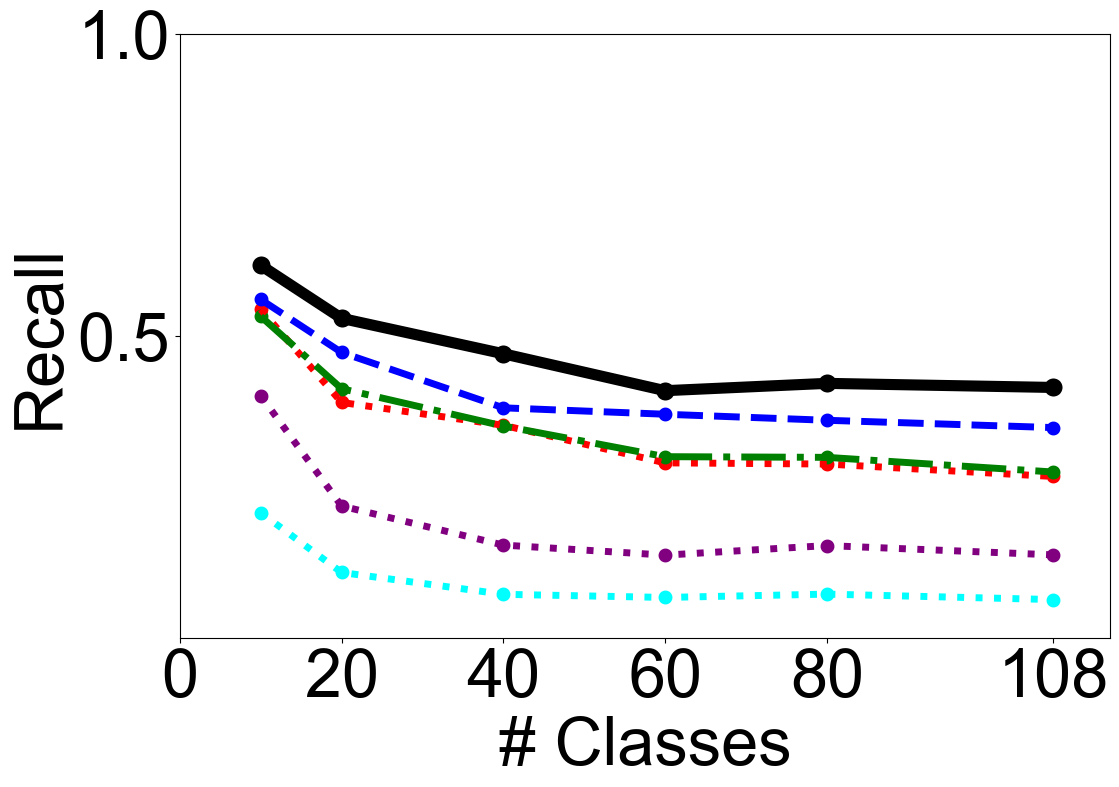}\label{fig:numberclasses_precision}}}
    \subfloat[\footnotesize{(d)} 
    ]{{\includegraphics[width=.5\columnwidth]{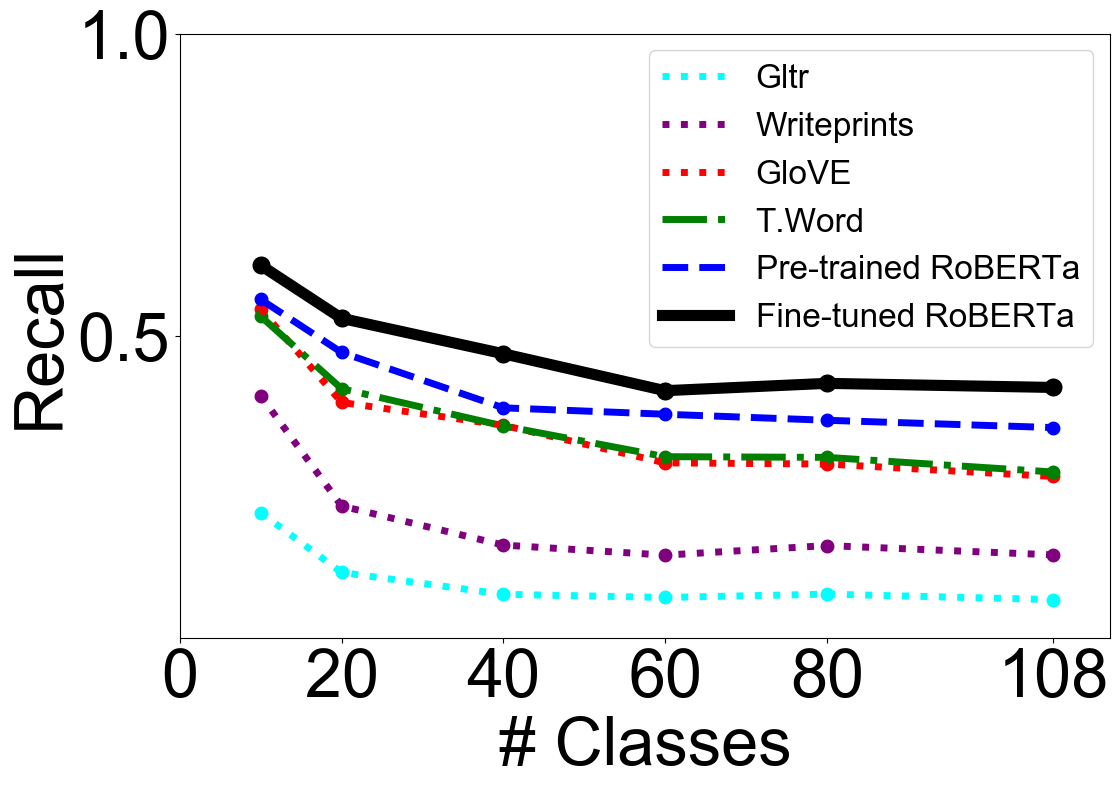}}\label{fig:numberclasses_recall}}
    \vspace{-2mm}
    \caption{\footnotesize{A comparison between the performances of various  embeddings by varying problem parameters. (a) Precision and (b) Recall with the varying training set size. (a) Precision and (b) Recall with the varying number of classes. Overall,  fine-tuned RoBERTa   outperforms all the baselines.}}
    \label{fig:all_train_var}
\end{figure*}

\subsubsection{Feature extraction}
Besides the feature extraction methods we mentioned in the Methods section of the main text, we also experimented with two more methods: \\
1. \textbf{Fine-tuned GLTR:}  Using the training set, we fine-tuned separate BERT and GPT2 models for each class for the task of mask completion. All models were then used for extracting GLTR word likelihood features for the complete training and test sets. Subsequently, the training representations were then fed to a sequential neural classifier like Bi-LSTM. The intuition was that word likelihoods extracted using the class's fine-tuned GLTR model would be high for synthetic text generated for the class's language model. For example, a r/wallstreetbets fine-tuned GLTR feature extractor would extract higher word likelihoods, as compared to other GLTR models, from  a synthetic comment generated by the language model fine-tuned on r/wallstreetbets. However, for an extremely small setting of 5 classes,  we only obtained a  precision and recall of around 33\% each. Due to the  (a) expensive cost of fine-tuning 108 BERT and GPT2 models, (b) expensive cost of extracting GLTR features for a dataset of 100k examples, (c) extremely  poor results of the approach on a small setting, we did not continue with experimenting the model in a larger setting.\\
2. \textbf{Trainable word embeddings} - We tokenized and represented each comment using an allocated set of integers based on the vocabulary of the training set. Then, similar to the GloVE feature extraction, we passed the representation into a trainable word embedding layer, followed by MLP or CNN. 
The results for these  were slightly worse than that for the GloVE features. 
We report the results in  Table \ref{tab:add_results}.

\begin{figure}[!t]
\captionsetup[subfigure]{labelformat=empty}
    \centering
    \subfloat[\footnotesize{(a) Micro} ]{{\includegraphics[width=0.5\columnwidth]{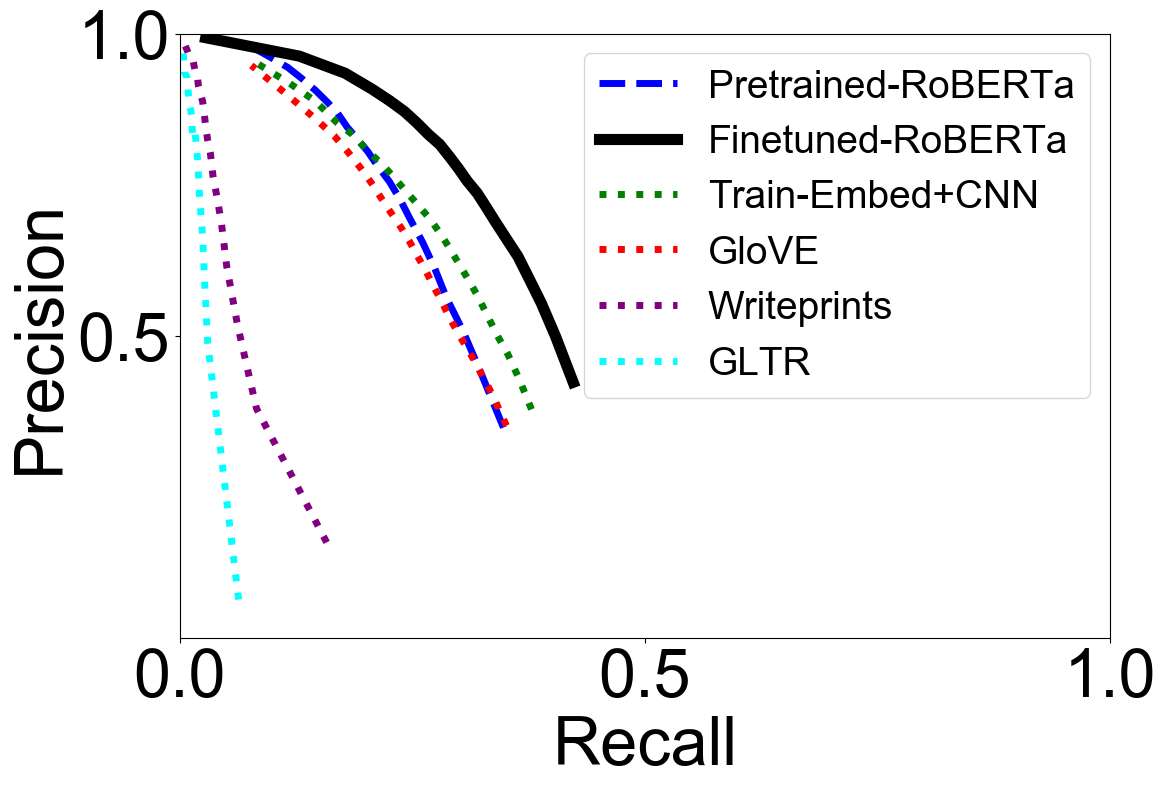}\label{fig:micro}}}
    \subfloat[\footnotesize{(b) Macro} ]{{\includegraphics[width=0.5\columnwidth]{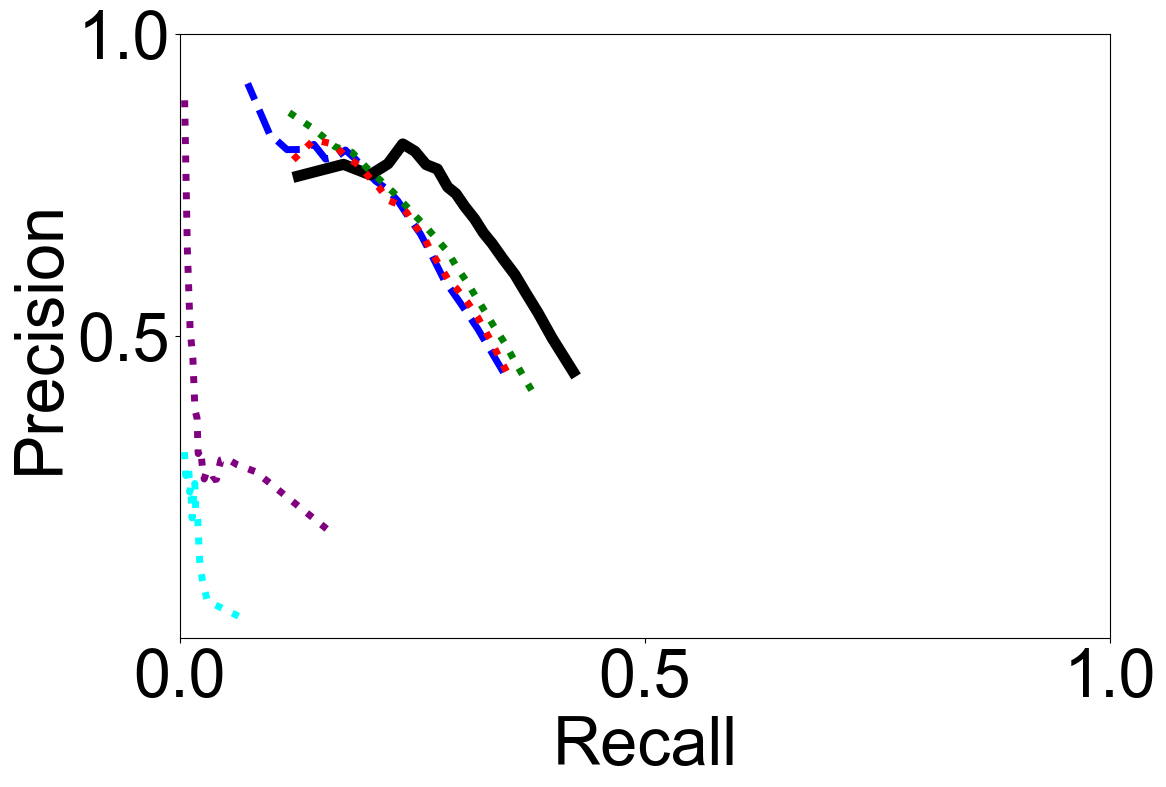}\label{fig:macro}}}
    \caption{\footnotesize{(a) Micro and (b) Macro \textit{precision-recall trade-off} by varying the gap statistic threshold. }}
    \label{fig:all_pr_tradeoff}
\end{figure}

Additionally, for all the feature representations mentioned in the main text, we tested SVM and  decision tree. Overall, they were outperformed by the other classifiers with one notable exception. For Writeprints features, SVM showed the best results of precision and recall of 19.3\% and 13.3\% respectively. 
We report the results in Table \ref{tab:add_results}.


\subsection{Other baselines}
\textbf{Precision-recall trade-off.}  For the precision-recall trade-off in the results section of the main text, we presented a comparison with additional baselines in Figure \ref{fig:all_pr_tradeoff}. Fine-tuned RoBERTa  performs the best among all methods for both micro and macro precision-recall trade-offs. They are followed in a decreasing order by the pre-trained RoBERTa embeddings, the trainable word embeddings, the GloVE word embeddings, Writeprints, and GLTR  respectively.\\
\textbf{Training set size and classes.} We report the comparison with all other baselines for the varying training size and the number of classes in Figure  \ref{fig:all_train_var}. Similar to the precision-recall trade-off, the fine-tuned RoBERTa embeddings perform the best, followed by pre-trained RoBERTa embeddings, GloVE word embeddings, trainable word embeddings, Writeprints, and GLTR.



\end{document}